\documentclass[10pt,twocolumn,letterpaper]{article}

\usepackage{cvpr}      

\usepackage[percent]{overpic}
\usepackage{amsmath}
\usepackage{multirow}
\usepackage{booktabs}
\usepackage[table,xcdraw,svgnames]{xcolor}
\usepackage{makecell}

\definecolor{cvprblue}{rgb}{0.21,0.49,0.74}
\usepackage{hyperref}
\hypersetup{
    colorlinks=true,      
    linkcolor=black,      
    citecolor=black,      
    urlcolor=blue,        
    pdfborder={0 0 0},    
}

\usepackage{amsmath}

\def\paperID{4902} 
\def\confName{CVPR}
\def\confYear{2026}

\title{ReHyAt: Recurrent Hybrid Attention for Video Diffusion Transformers}

\author{Mohsen Ghafoorian, Amirhossein Habibian \\
Qualcomm AI Research\thanks{Qualcomm AI Research is an initiative of Qualcomm Technologies, Inc.} \\
\texttt{\{mghafoor,ahabibia\}@qti.qualcomm.com} \\
}

\begin{document}
\maketitle
\begin{abstract}
Recent advances in video diffusion models have shifted towards transformer-based architectures, achieving state-of-the-art video generation but at the cost of quadratic attention complexity, which severely limits scalability for longer sequences. We introduce ReHyAt, a Recurrent Hybrid Attention mechanism that combines the fidelity of softmax attention with the efficiency of linear attention, enabling chunk-wise recurrent reformulation and constant memory usage. Unlike the concurrent linear-only SANA Video, ReHyAt’s hybrid design allows efficient distillation from existing softmax-based models, reducing the training cost by two orders of magnitude to $\sim$160 GPU hours, while being competitive in the quality. Our light-weight distillation and finetuning pipeline  provides a recipe that can be applied to future state-of-the-art bidirectional softmax-based models. Experiments on VBench and VBench-2.0, as well as a human preference study, demonstrate that ReHyAt achieves state-of-the-art video quality while reducing attention cost from quadratic to linear, unlocking practical scalability for long-duration and on-device video generation. Project page is available at \href{https://qualcomm-ai-research.github.io/rehyat}{https://qualcomm-ai-research.github.io/rehyat}.
\end{abstract}

\section{Introduction}
\begin{figure}[!th]
    \centering

    \begin{subfigure}[t]{0.49\linewidth}
        \centering
        \includegraphics[width=\linewidth,height=0.15\textheight]{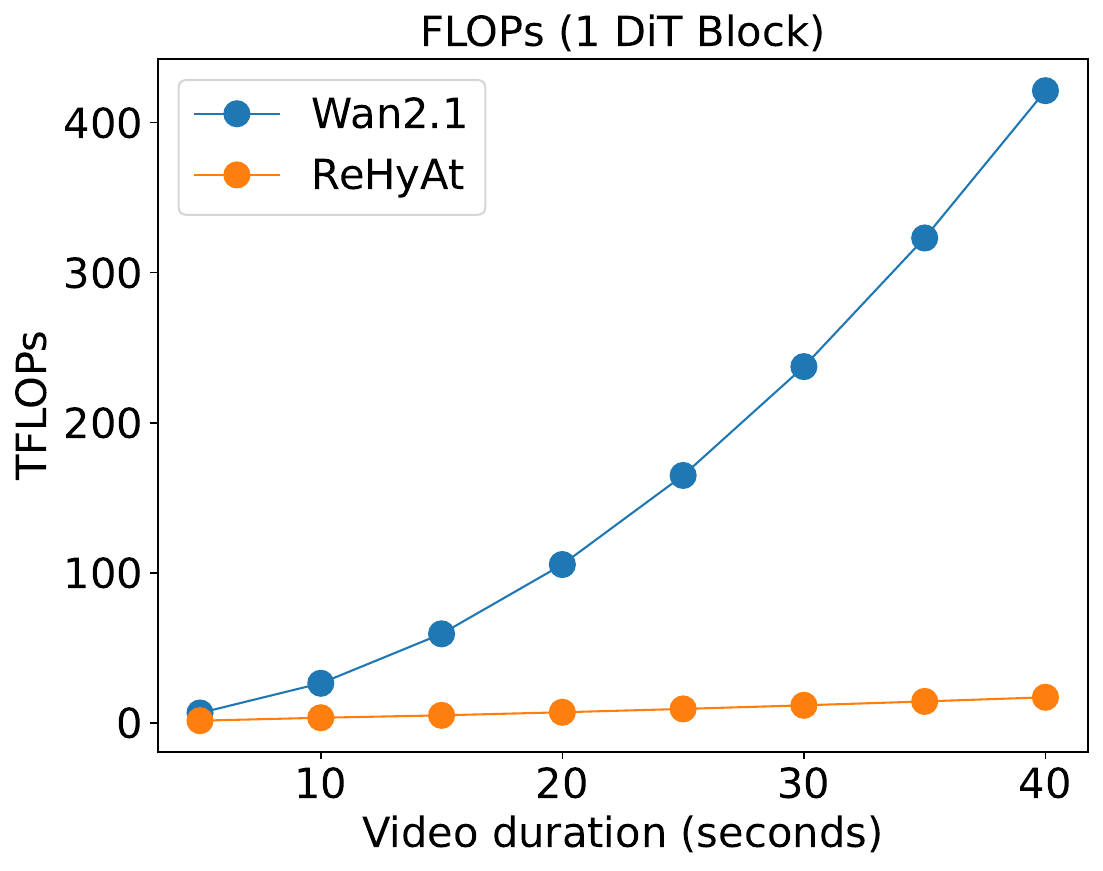}
    \end{subfigure}
    \hfill
    \begin{subfigure}[t]{0.49\linewidth}
        \centering
        \begin{overpic}[width=\linewidth,height=0.15\textheight]{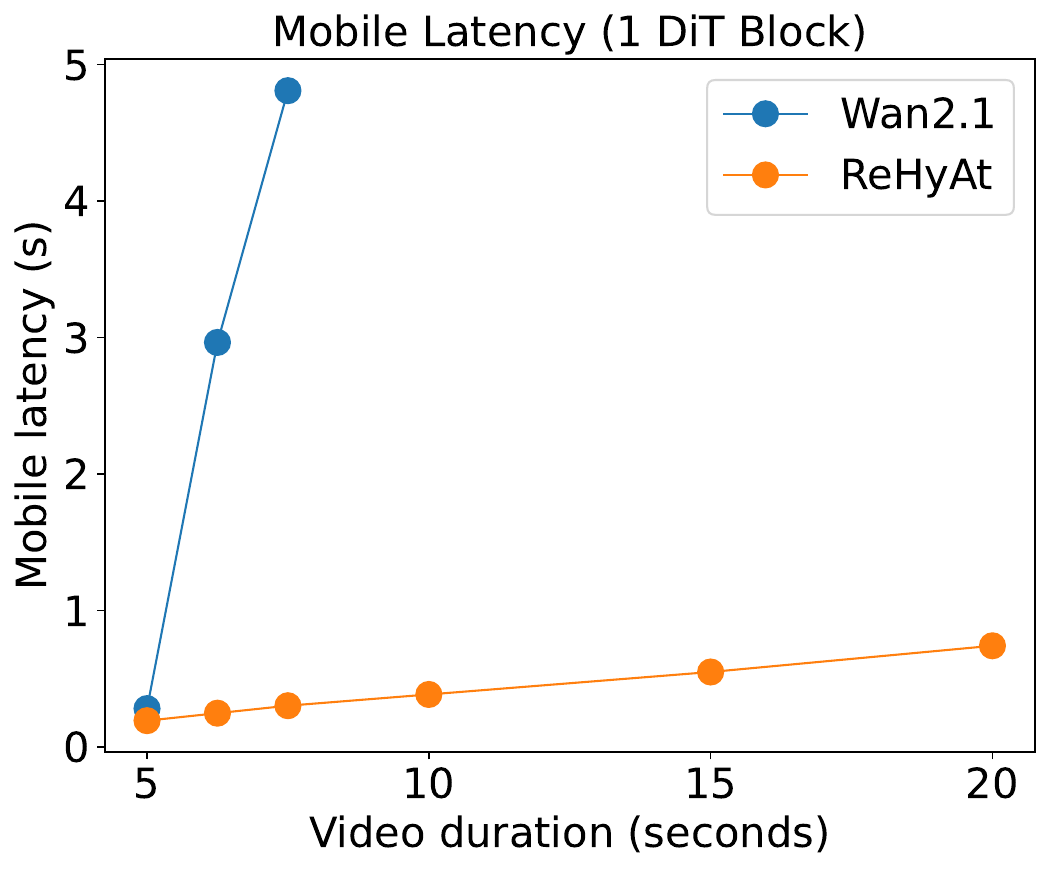}
            \put(30,71){\scriptsize \textbf{\textcolor[HTML]{1F77B4}{OOM}}}
        \end{overpic}
    \end{subfigure}

    \vspace{0.5em}
    \begin{subfigure}[t]{\linewidth}
        \centering
        \begin{minipage}[c]{0.06\linewidth}
            \centering
            \rotatebox{90}{\scriptsize \shortstack{Wan2.1 1.3B\\VBench: \textbf{83.1}}}
        \end{minipage}%
        \hfill
        \begin{minipage}[c]{0.92\linewidth}
            \centering
            \includegraphics[width=\linewidth]{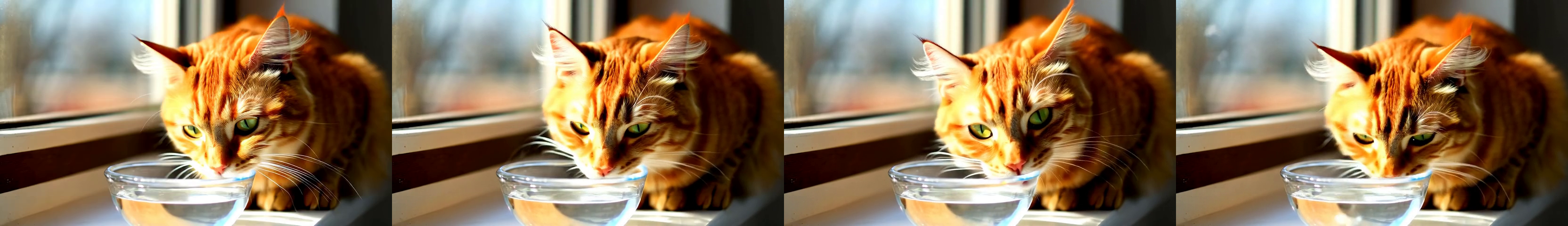}
        \end{minipage}
    \end{subfigure}

    \vspace{0.5em}
    \begin{subfigure}[t]{\linewidth}
        \centering
        \begin{minipage}[c]{0.06\linewidth}
            \centering
            \rotatebox{90}{\scriptsize \shortstack{20$\times$ReHyAt\\VBench: \textbf{83.4}}}
        \end{minipage}%
        \hfill
        \begin{minipage}[c]{0.92\linewidth}
            \centering
            \includegraphics[width=\linewidth]{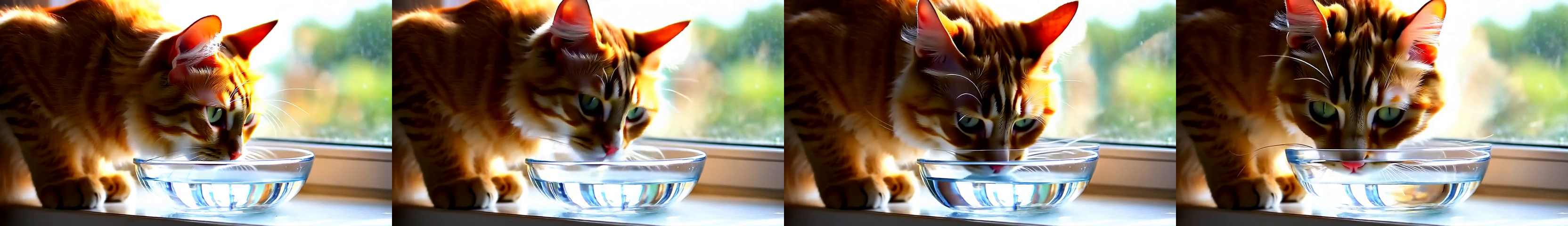}
        \end{minipage}
    \end{subfigure}

    \caption{A comparison of our proposed Recurrent Hybrid Attention model with Wan2.1 bidirectional full softmax attention. Top: Compute complexity increase with video duration growth (left: FLOPs, right: phone latency). Bottom: comparing our hybrid model (20$\times$ ReHyAt blocks) with original Wan2.1 1.3B, qualitatively and quantitatively. Prompt: ``A cat drinking water.''}
    \label{fig:teaser}
\end{figure} 
The ambition in generative video is shifting from producing short, visually striking clips to creating sustained, coherent sequences with rich dynamics and consistent subject identity. Diffusion-based models have become the method of choice for this goal due to their stability and controllability; however, the choice of backbone is decisive for scaling. While early video diffusion systems adapted U-Net architectures from images, they exhibited limited capacity to model long temporal structure and struggled to scale effectively to higher resolutions and durations. This has motivated a transition to Diffusion Transformers (DiTs)~\citep{peebles2023dit}, which process video as a sequence of spatiotemporal patches and furnish global context from the first layer. The resulting architectural shift underlies recent state-of-the-art systems (e.g., \textit{Wan2.1}~\citep{wan2025}, \textit{CogVideoX}~\citep{yang2025cogvideox}, \textit{HunyuanVideo}~\citep{tencent2025hunyuan}, \textit{PyramidalFlow}~\citep{liu2025pyramidalflow}, \textit{Open-Sora Plan}~\citep{lin2024opensora}), and has been documented by recent surveys as the prevailing trend in video generation~\citep{wang2025survey,melnik2024survey}.

This progress comes with a nontrivial systems cost: the self-attention term scales quadratically with sequence length, $\mathcal{O}(N^2 d)$ in time and $\mathcal{O}(N^2)$ in memory, where $N$ is the number of tokens and $d$ the hidden dimension~\citep{vaswani2017attention,rabe2021memory}. In video, $N$ is the product of temporal length and spatial patch count, so even moderate resolutions and durations yield token counts in the tens of thousands. In practice, the attention subroutine consumes the majority of compute in DiT blocks, and memory pressure grows rapidly with longer contexts. Kernel- and IO-aware implementations such as FlashAttention~\citep{dao2022flashattention} reduce constants but do not alter the $N^2$ dependence, leaving training and inference constrained when targeting higher resolutions, extended durations, or multi-shot compositions. As a direct consequence, producing videos beyond roughly 10 seconds remains difficult within typical GPU memory and latency budgets, while edge devices such as mobile phones even struggle to generate more than a few seconds of videos.

Linear attention~\citep{katharopoulos2020linear} offers a compelling alternative to full softmax attention by reducing complexity from quadratic to linear and enabling constant memory when reformulated as an RNN. This property makes it particularly attractive for generating arbitrarily long videos, where memory growth is a critical bottleneck. Beyond efficiency, the recurrent formulation of linear attention allows chunk-wise processing, which aligns naturally with sequential video generation. These advantages have motivated recent efforts to explore linear and hybrid attention mechanisms in video diffusion models~\citep{chen2025sana, ghafoorian2025attention}. 

However, linear attention introduces a significant trade-off: its kernel-based similarity function lacks the expressiveness of the exponential kernel used in softmax attention. This gap manifests in reduced activation diversity and weaker modeling of fine-grained dependencies~\citep{zhang2024hedgehog}, often requiring extensive retraining to achieve acceptable quality~\citep{chen2025sana, ghafoorian2025attention}. Hybrid approaches that combine linear and softmax attention have emerged as a potential solution~\citep{ghafoorian2025attention}, but existing designs remain quadratic in complexity and cannot be reformulated as RNNs, leaving the scalability challenge unresolved. In other words, while these methods improve quality over purely linear attention, they fail to deliver the memory and compute benefits necessary for long-duration video generation.

Meanwhile, the most powerful video diffusion models today are trained with bidirectional full softmax attention using massive compute and data resources. Re-training such models with alternative attention mechanisms from scratch is prohibitively expensive and impractical for most research and production settings. This observation motivates a different strategy: rather than building efficient models from the ground up, can we distill these high-quality, compute-heavy models into a recurrent form that preserves fidelity while dramatically reducing resource requirements? Achieving this would unlock practical scalability for video diffusion, not neglecting the substantial progress made by state-of-the-art architectures.

In this paper, we address this challenge by introducing \textbf{ReHyAt}, a recurrent hybrid attention mechanism tailored for video diffusion. Our key insight is that preserving softmax attention for a small subset of tokens—those most critical for modeling local dependencies—while applying linear attention globally enables modeling long-range and high fidelity local dependencies while ensuring linear efficiency. We propose a temporally chunked hybrid attention design with overlapping chunks to maintain temporal coherence, and show that this formulation can be reformulated into a chunk-wise RNN with constant memory complexity. Furthermore, we leverage a two-stage training pipeline—attention distillation from a bidirectional softmax teacher followed by lightweight fine-tuning—that achieves SOTA results within fewer than 200 GPU-hours. We validate our approach by transforming \textit{Wan2.1} into its recurrent hybrid counterpart and evaluate on VBench~\citep{huang2023vbench}, VBench2.0~\citep{zheng2025vbench}, and a human preference study, demonstrating that ReHyAt delivers near state-of-the-art quality with dramatically reduced compute. Fig~\ref{fig:teaser} demonstrates some of the aspects discussed above.

\noindent Our main contributions are as follows:
\begin{itemize}
    \item We propose \textit{ReHyAt}, a novel temporally chunked hybrid attention mechanism that combines local softmax attention with global linear attention. This design preserves high-fidelity modeling of critical dependencies within and across adjacent frames while reducing overall complexity to linear time.
    \item We derive a chunk-wise \textit{recurrent} reformulation of ReHyAt, computationally enabling generation of arbitrarily long videos with constant memory usage and efficient inference.
    \item Through extensive empirical evaluations and ablation studies, we show that a state-of-the-art bidirectional Softmax attention video diffusion model can be transformed into a chunk-wise recurrent model, only within a few hundred GPU-hours, with negligible impact on the quality.    
\end{itemize}

\section{Related Work}
\textbf{Efficient Attention.}
Several approaches aim to reduce the quadratic complexity of self-attention across domains: for vision tasks (e.g., EfficientViT~\citep{cai2022efficientvit}, PADRe~\citep{letourneaupadre}, Performer~\citep{choromanski2020performer}, Linformer~\citep{wang2020linformer}), image generation (e.g., SANA~\citep{li2023sana}, LinGen~\citep{wang2025lingen}, Grafting~\citep{Grafting}), and language modeling~\citep{mercat2024linearizing, wang2024mamba, yang2024parallelizing, chen2024hedgehog, zhang2024lolcats}. These works show the feasibility of sub-quadratic attention but often require heavy retraining or training from scratch (e.g., SANA~\citep{li2023sana}). In contrast, we focus on lightweight distillation and fine-tuning of pre-trained softmax-based models into an efficient hybrid attention design tailored for video diffusion under modest compute budgets.
Linear recurrent models such as SSM and RWKV~\citep{fei2024diffusion, fei2024dimba, wang2024mamba, yao2025diffusion, zhu2025dig} have emerged as alternatives to self-attention for long sequences. However, architectural differences from transformers make distilling DiT weights into these models costly. Our approach preserves the original block structure, enabling effective distillation with minimal training. Finally, as noted in Katharopoulos et al.~\citep{katharopoulos2020linear}, causal linear attention can be reformulated as an RNN during inference—a property we leverage for efficient long video generation.

\noindent \textbf{Video Diffusion Models.}
Recent large-scale systems such as CogVideoX~\citep{yang2025cogvideox}, Open-Sora Plan~\citep{lin2024opensora}, PyramidalFlow~\citep{liu2025pyramidalflow}, LTX-video~\citep{hacohen2024ltx}, and Wan2.1~\citep{wan2025} have significantly advanced video generation quality and scalability, but at substantial compute and memory cost. Mobile/PC-oriented designs like Mobile Video Diffusion~\citep{yahia2024mobile}, MoViE~\citep{karjauv2024movie}, SnapGen-V~\citep{wu2025snapgen}, AMD-HummingBird~\citep{isobe2025amd}, On-device Sora~\citep{kim2025device}, MobileVDiT~\citep{wu2025taming}, and NeoDragon~\citep{karnewar2025neodragon} aim for lightweight deployment, yet most remain non-DiT-based or still rely on full quadratic attention, limiting scalability for long-duration videos.

\noindent \textbf{Video Diffusion Models with Efficient Attention.}
Prior work has explored accelerating video generation through token merging~\citep{bolya2023token,kahatapitiya2024object,evdit}, token downsampling~\citep{crowson2024scalable,peruzzo2025adaptor}, attention tiling~\citep{evdit,zhang2025fast}, and sparsity~\citep{li2025compact,zhang2025faster}. Tiling and sparsity-based approaches, in particular, gain efficiency by discarding attention for most tokens. In contrast, our hybrid attention design attends to the full token set, combining linear attention for long-range dependencies with softmax attention for local, high-fidelity interactions. M4V~\citep{huang2025m4v} accelerates video DiTs by distilling them into Mamba blocks. Despite our simpler block structure and lightweight training, we outperform M4V in both quality and efficiency.

Recently, Attention Surgery~\citep{ghafoorian2025attention} proposed a temporally uniform hybrid attention method with reasonable quality but retained quadratic complexity. Our approach introduces a temporally non-uniform hybrid arrangement, enabling uneven treatment of token dependencies and a better inductive bias for video generation. It achieves linear complexity and can be reformulated as a memory-efficient RNN, supporting on-device execution and scalable long video generation.

Finally, concurrent to our work, SANA-Video~\citep{chen2025sana} introduced a video diffusion model incorporating linear attention. In contrast, our method offers a hybrid approach combining the computational efficiency of linear attention for long-range dependencies with the accuracy of softmax attention for modeling highly co-dependent adjacent tokens. Furthermore, unlike SANA-Video, our method sets up a distillation process from a SOTA bidirectional full softmax attention model, making training extremely efficient: we obtain our model in $\sim$160 GPU-hours—\textit{two orders of magnitude more efficient than SANA-Video}. This work therefore provides a low-cost recipe to transform costly Softmax attention SOTA models into efficient RNNs, laying the groundwork for long video generation and on-device execution.

\section{Methods: ReHyAt}
\subsection{Preliminaries: Linear Attention}
Let \( x \in \mathbb{R}^{N \times D} \) denote a sequence of \( N \) tokens, each represented by a \( D \)-dimensional feature vector. At the \( l \)-th transformer layer, the block is formulated as:
\begin{equation}
    T_l(x) = f_l\big(A_l(x) + x\big),
\end{equation}
where \( f_l(\cdot) \) applies a token-wise transformation, typically a lightweight feed-forward network, and \( A_l(\cdot) \) represents the self-attention operator—the component responsible for cross-token interaction. The standard attention mechanism is given by:
\begin{equation}
A_l(x) = y = \text{softmax}\!\left(\frac{q k^\top}{\sqrt{D}}\right)v,
\label{eq:softmax}
\end{equation}
where queries, keys, and values are computed as linear projections:
\[
q = x w_q,\; k = x w_k,\; v = x w_v,
\]
with learnable weights \( w_q, w_k, w_v \in \mathbb{R}^{D \times D} \) .

The softmax attention for token \( i \) can be expressed as:
\begin{equation}
y_i = \frac{\sum_{j=1}^{N} \text{sim}(q_i, k_j)\, v_j}{\sum_{j=1}^{N} \text{sim}(q_i, k_j)}.
\end{equation}

Applying the kernel trick, the similarity function can be generalized from
\(\text{sim}(q_i, k_j) = e^{q_i k_j^\top}\) (recovering the original softmax)  
to
\(\text{sim}(q_i, k_j) = \phi(q_i)\phi(k_j)^\top\),  
yielding:
\begin{equation}
y_i = \frac{\phi(q_i)\sum_{j=1}^{N}\phi(k_j)\, v_j^\top}{\phi(q_i)\sum_{j=1}^{N}\phi(k_j)}.
\end{equation}

Crucially, the terms \(\sum_{j=1}^{N}\phi(k_j)\, v_j^\top\) and \(\sum_{j=1}^{N}\phi(k_j)\) do not depend on \( i \), enabling precomputation and caching for linear-time complexity. The mapping \(\phi(\cdot)\) must be non-negative; the original work by \citet{katharopoulos2020linear} proposes \(\phi(x) = 1 + \text{elu}(x)\). However, this substitution introduces a notable gap in expressiveness compared to the exponential kernel, often requiring substantial retraining or resulting in degraded performance relative to softmax attention.

\subsection{Hybrid Attention Formulation}
Before introducing the formal expression, we note that the hybrid attention mechanism combines contributions from both softmax attention (for local, high-fidelity dependencies) and linear attention (for global, efficient interactions), and normalizes them jointly.

For the latent \( x \in \mathbb{R}^{N \times D} \), assume the $N$ tokens are flattened from a latent tensor of shape $(T, H, W, D)$, where $N=THW$. To overcome the limitations of purely linear attention in video diffusion models, we incorporate a \emph{hybrid attention} mechanism that combines softmax-based and kernelized linear attention formulations. Now consider a chunk of $T_c$ temporal slices from the latent, represented as $X_t \in \mathbb{R}^{N'\times D}$, where $N'=T_cHW$. Here we have introduced the chunk-indexed reshaped notation $X \in \mathbb{R}^{T'\times N' \times D}$, with $T'=N/N'$ representing the number of chunks, to avoid confusion with single token indexing e.g. $x_i$.
Following the same notation, we have $Q_t \in \mathbb{R}^{N'\times D}$, and $\phi_q(Q_t) \in \mathbb{R}^{N'\times D'}$. Then for the hybrid attention of tokens in chunk $t$, we partition the total tokens $\mathcal{T}=\{1,2...,N\}$ to attend to, into softmax attention tokens $\mathcal{T}_t^\text{S}$ and linear attention tokens $\mathcal{T}_t^\text{L}$. More specifically, the hybrid attention output for token chunk $t$, $\hat{y}_t \in \mathbb{R}^{N'\times D}$ constitutes of softmax attention and its normalizer $a_t^\text{S} \in \mathbb{R}^{N'\times D} $ and $n_t^\text{S} \in \mathbb{R}^{N' \times 1}$ as well as linear attention and its normalization term $a_t^\text{L} \in \mathbb{R}^{N'\times D}$, $n_t^\text{L} \in \mathbb{R}^{N' \times 1}$, formulated as below:
\begin{equation}
    \hat{y}_t = \frac{a_t^\text{S} + a_t^\text{L}}{n_t^\text{S} + n_t^\text{L}},
\end{equation}
\begin{align}
a_t^\text{S} &= \sum_{j\in \mathcal{T}_t^\text{S}} \exp(Q_t k_j^\top / \sqrt{D} - c_t) v_j, \\
a_t^\text{L} &= \phi_q(Q_t) \Big( \sum_{j \in \mathcal{T}_t^\text{L}} \phi_k (k_j)\, v_j^\top \Big), \\
n_t^\text{S} &= \sum_{j \in \mathcal{T}_t^\text{S}} \exp(Q_t k_j^\top / \sqrt{D} - c_t), \\
n_t^\text{L} &= \phi_q(Q_t) \Big( \sum_{j \in \mathcal{T}_t^\text{L}} \phi_k (k_j) \Big),
\label{eq:hybrid}
\end{align}
where \( c_t \) is a stabilizing constant (typically the maximum exponent), and \( \phi_q(\cdot) \) and \( \phi_k(\cdot) \) denote the kernel feature maps for the linear component for queries and keys.

\begin{figure}[t]
\centering
\begin{overpic}[width=\columnwidth]{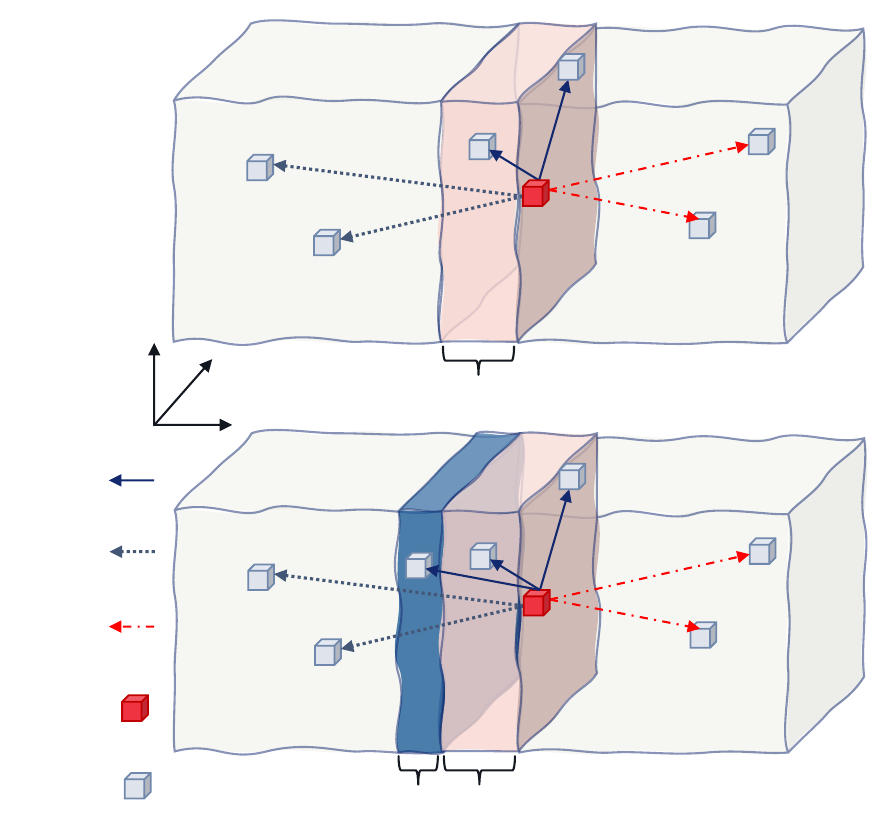}

  \put(14.5,54){\small$Y$}
  \put(24,51){\small$X$}
  \put(26.5,45){\small$T$}
  
  \put(-8,3){\parbox[c]{2cm}{\centering \scriptsize Source \\ Tokens $j$ ($k,v$)}}
  \put(-8,11){\parbox[c]{2cm}{\centering \scriptsize Target \\ Token $i$ ($q$)}}
  \put(-8,30){\parbox[c]{2cm}{\centering \scriptsize Causal Linear \\ Attention}}`
  \put(-8,21){\parbox[c]{2cm}{\centering \scriptsize Non-causal \\ Linear \\ Attention}}
  \put(-8,38){\parbox[c]{2cm}{\centering \scriptsize Softmax \\ Attention}}
  
  \put(53,49){\footnotesize $T_c$}
  \put(53.5,2){\footnotesize $T_c$}
  \put(46.5,2){\footnotesize $T_o$}
  
\end{overpic}
\vspace{-15pt}
\caption{Overview of the temporally chunked hybrid attention arrangement without (top) and with chunk overlap (bottom).}

\label{fig:method}

\end{figure}
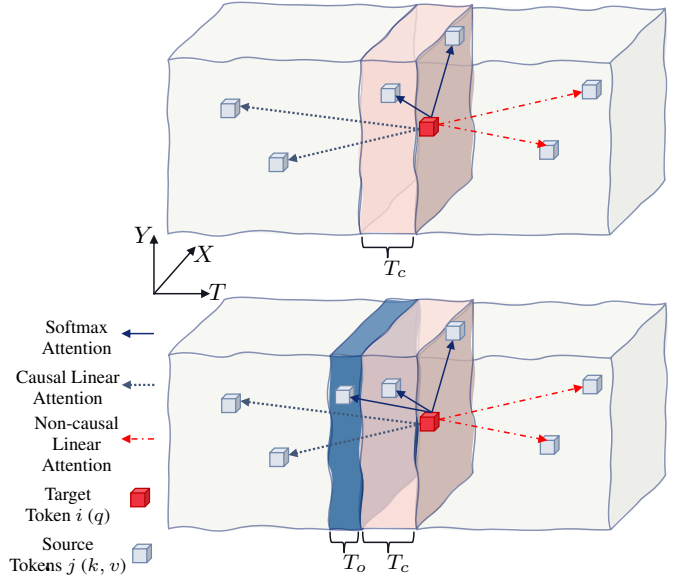

Here, we propose the following specification for the partitioning of the tokens sets:
\begin{align}
    \mathcal{T}_t^\text{S} &= \{j \ | \ tN'\leq j < (t+1)N'\}, \\
    \mathcal{T}_t^\text{L} &= \mathcal{T} - \mathcal{T}_t^\text{S}
\end{align}
 See the top graph in Fig~\ref{fig:method}. This means that the computation of attention is effectively broken into temporal chunks of $T_c$ slices, where the tokens within each slice more accurately attend to each other with the Softmax attention, and with linear attention to all the other tokens.

\noindent \textbf{Overlapping Chunks}. We observe that the non-overlapping chunking mechanism defined above, together with the lower fidelity dependency modeling of linear attention, can result into episodic incoherence in motion or appearance between the frames transitioning from one latent chunk to next. To mitigate this, we propose to arrange overlapping chunks for softmax attention, enabling a more accurate softmax-attention-based message passing between the chunks. More specifically, for generating attention output for a chunk given chunk of $T_c$ slices (i.e. applying this slicing to queries), the keys and values representing the tokens to attend to, are sliced by $T_c + T_o$ temporal slices instead, where $T_o$ represents the overlap size. To arrange this, one needs to reformulate $\mathcal{T}_t^{S}$ and $\mathcal{T}_t^{L}$ as:
\begin{align}
    \mathcal{T}_t^\text{S} &= \{j \ | \max(tN' - T_oHW, 0) \leq j < (t+1)N'\} \nonumber\\ 
    \mathcal{T}_t^\text{L} &= \mathcal{T} - \mathcal{T}_t^\text{S}
\end{align}
The bottom subgraph in Fig~\ref{fig:method} illustrates this.

\textbf{Characterization of $\phi$.}  
Similar to~\cite{ghafoorian2025attention}, to enhance the expressiveness of linear attention, we define distinct learnable feature maps $\phi_q, \phi_k : \mathbb{R}^D \rightarrow \mathbb{R}^{D'}$. Each map first applies a lightweight per-head embedding network (implemented as grouped $1\times1$ convs with non-linear activations) to produce an intermediate representation, which is then split into $P$ equal parts. Each part is raised to a different polynomial degree $1$ to $P$, and concatenated along the feature dimension. Formally, for an input $x \in \mathbb{R}^D$, we define:
\[
\phi(x) = [\;(\psi_1(x))^1, (\psi_2(x))^2, \dots, (\psi_P(x))^P\;]^\top \in \mathbb{R}^{D'},
\]
where $\psi_i(\cdot)$ denotes the $i$-th learnable embedding slice produced by the shared embedding network. This polynomial expansion allows $\phi_q(q_i)\phi_k(k_j)^\top$ to approximate the large dynamic range of the exponential kernel $e^{q_i k_j^\top}$ more accurately than fixed ELU-based mappings.

\subsection{Recurrent HyAt}
Linear attention, once causal has the advantage that can be reformulated to RNNs. In this section we show how our hybrid arrangement, unlike ~\cite{ghafoorian2025attention}, can be reformulated as an RNN. For this to be feasible, we first need to make it causal. To achieve this, it is sufficient to reformulate $\mathcal{T}_t^\text{L}$ as follows:
\begin{align}
    \mathcal{T}_t^\text{L} &= \{j \ | j < \max(tN' - T_oHW, 0)\} \\
    \mathcal{T}_t^\text{S} &= \{j \ | \max(tN' - T_oHW, 0) \leq j < (t+1)N'\} \nonumber
\end{align}
In Fig.~\ref{fig:method}, this is equivalent to the bottom graph where the specified non-causal linear attention is removed.
Now thanks to the temporal decoupling of $\mathcal{T}_t^\text{S}$ and $\mathcal{T}_t^\text{L}$, we can define a chunk-wise RNN, where the model generates the latents chunk-by-chunk for $T_c$ temporal slices at a time. Let $s_t \in \mathbb{R}^{D'\times D} $ and $z_t \in \mathbb{R}^{D' \times 1}$ represent the state variables for the linear attention and its normalizer, $t$-th chunk. Then we have:
\begin{align}
    s_0 &= 0 \\
    z_0 &= 0 \\
    y_t &= \frac{a_t^\text{S} + \phi_q(Q_t)\, s_t}{n_t^\text{S} + \phi_q(Q_t)\, z_t}\\
    s_{t+1} &= s_t + \sum_{j \in \mathcal{T}_t^\text{L}}{\phi_k(k_j)\, v_j^\top} \\
    z_{t+1} &= z_t + \sum_{j \in \mathcal{T}_t^\text{L}}{\phi_k(k_j)}
\end{align}
Three points to note: (1) Softmax attention within each chunk need not be causal because sampling proceeds chunk-by-chunk, i.e. our method generates the latents for a full chunk at once. (2) The model training doesn't have to be done in the RNN form as introduced above. One can train the model in the causal non-recurrent form and then rearrange the trained model to RNN at the sampling time. (3) The computational complexity remains $\mathcal{O}(N)$ with the length of the generated video, while the memory complexity remains constant irrespective of the video duration. 
\subsection{Two-stage Training}
\label{sec:attention_surgery}
Given the enormous compute/data requirements for obtaining SOTA video diffusion models, our proposed method is instead centered around efficiently distilling existing bidirectional full softmax attention, e.g. Wan2.1, into our proposed RNN formulation. To achieve this, we propose a two-stage process: attention distillation and lightweight finetuning that we expand in the following. Thanks to this specific method design, we obtain a recurrent video diffusion model with competitive quality, within less than 200 GPU-hours.
\subsubsection{Attention Distillation}
We first distill a bidirectional full softmax teacher model into a causal hybrid attention student model. During this stage, each block is trained independently and the only learnable parameters are $\phi_q$ and $\phi_k$ per block, so as to let $\phi$ parameters to enable linear attention to approximate the corresponding softmax dependencies. This distillation setup doesn't require any prompt/video pairs for the training; the student model is trained to match the teacher activations for different prompts, noise samples and denoising iterations. The following equation formalizes this:
\begin{equation}
\boldsymbol{\phi}_l = \boldsymbol{\phi}_l - \eta\, \nabla_{\boldsymbol{\phi}_l} \Bigg(
\mathbb{E}_{\substack{\epsilon \in \mathcal{N} \\ p \in \mathcal{P} \\ i \in \mathcal{S}}}
\big|y^{(l, \epsilon, p, i)} - \hat{y}^{(l, \epsilon, p, i)}\big|
\Bigg),
\end{equation}
where $\boldsymbol{\phi}_l$ is $(\phi_q, \phi_k)$ for the the $l$-th block, $\mathcal{N}$ the noise sampling distribution, $\mathcal{P}$ the distribution of textual prompts, $\mathcal{S}$ the set of denoising steps, $y^{(l, \epsilon, p, i)}$ the output of the bidirectional softmax teacher on block $l$, for prompt $p$, sampling noise $\epsilon$, and denoising step $i$, and $\hat{y}^{(l, \epsilon, p, i)}$ the same trajectory point for the ReHyAt student model.

\subsubsection{Lightweight Fine-tuning}
After the pretraining distillation stage making the block attentions recurrent hybrid, we have obtained the $\phi_q$s and $\phi_k$s per block. However, while the pretraining distillation helps preserve the general structure of the scenes, the details will be far from perfect, specifically on the transition smoothness between chunks, as the blocks are pretrained in isolation. Now fine-tuning the whole DiT model on a modest set of prompt/video pairs, for a small number of iterations (e.g. 1k) recovers the lost generation quality. This is done by optimizing the normal flow-matching objective~\cite{lipman2022flow}. 
\section{Experimental Setup}

\subsection{Evaluation of generation quality}
We evaluate ReHyAt by distilling and fine-tuning Wan2.1 1.3B model~\citep{wan2025}, a widely used efficient SOTA model. For SOTA comparisons, we generate videos at the original Wan resolution and length (\(81\times480\times832\)) using the full set of extended prompts from VBench~\citep{huang2023vbench} and VBench-2.0 ~\citep{zheng2025vbench}.

In addition to quantitative evaluation, we conduct a blinded human preference study to assess visual qualities and prompt alignment. We randomly select 50 prompts from VBench and present participants with paired videos, asking them to choose their preferred video or indicate no significant difference. The order of paired videos randomly change per prompt to avoid any potential biases. In total, we collect 500 paired comparisons.

To enable large-scale ablation studies, we train and evaluate our model variants at a lower spatial resolution of \(320 \times 480\) per frame. For all evaluations, we use the model snapshot at the 1000th fine-tuning iteration.

\subsection{Assessment of compute complexity}
\noindent \textbf{FLOPs Analysis}.
We analyze the number of floating point operations in the proposed ReHyAT method and compare it against flash attention, and other alternatives on the original 5-second Wan video generation setup, as well as analyzing the DiT blocks’ compute growth as we increase the length of generated videos. For this, we use the DeepSpeed library to measure the complexities.

\noindent \textbf{On-mobile Measurements}.
 A valuable advantage of the proposed recurrent hybrid attention method is that it computationally enables the generation of longer videos on edge devices such as mobile phones, thanks to lower compute burden, and most importantly, due to significant reduction in peak memory consumption. We port the transformed original WAN model with flash attention blocks as well as the transformed ReHyAt modules to Qualcomm AI Runtime (QNN) and profile run-time metrics such as latency, memory read, and memory write on a Snapdragon8-Gen4 SoC. For the on-device measurements we report the metrics on 320$\times$480 frame resolution with the original 5 seconds WAN video length, as well as longer video durations. 

\subsection{Training specification}

\textbf{Datasets.}  
For fine-tuning low-resolution models, we use a 350K subset of the video dataset from Open-Sora Plan~\citep{lin2024opensora}. For high-resolution fine-tuning, we use 22K synthetic video samples generated by Wan2.1 14B, with prompts drawn from the same source as used for the low-resolution dataset.

\noindent \textbf{Model Hyperparameters.}
We experiment with converting different numbers of transformer blocks to recurrent hybrid attention: 15, 20, and 25 out of the 30 blocks in Wan2.1 1.3B. For the hybrid blocks, we explore hybridization with various chunk sizes ($T_c \in \{1, 2, 3, 5, 7\}$) as well as different options for overlap size ($T_o \in \{0, 1, 2, 3\}$). Empirical analysis of the impact of $\phi_k$ and $\phi_q$ transformation complexity on generation quality shows that a lightweight 2-layer MLP with degree-2 polynomial features is sufficient. This configuration adds approximately 2.4M parameters per converted block. Additional details are provided in the appendix.

\begin{table}[t]
\centering
\footnotesize
\setlength{\tabcolsep}{4pt}
\renewcommand{\arraystretch}{1.05}
\resizebox{\linewidth}{!}{%
\begin{tabular}{lccc}
\hline
\textbf{Models with 2B–5B parameters} & Total$\uparrow$ & Quality$\uparrow$ & Semantic$\uparrow$ \\
\hline
Open-Sora Plan V1.3~\citep{lin2024opensora} & 77.23 & 80.14 & 65.62 \\
CogVideoX 5B~\citep{yang2025cogvideox} & 81.91 & 83.05 & 77.33 \\
CogVideoX1.5 5B~\citep{yang2025cogvideox} & 82.01 & 82.72 & 79.17 \\
\hline
\textbf{Models up to 2B parameters} & & & \\
\hline
Open-Sora V1.2~\cite{zheng2024open} & 79.76 & 81.35 & 73.39 \\
LTX-Video~\citep{hacohen2024ltx} & 80.00 & 82.30 & 70.79 \\
SnapGenV~\citep{wu2025snapgen} & 81.14 & 83.47 & 71.84 \\
Hummingbird 16frame~\citep{isobe2025amd} & 81.35 & 83.73 & 71.84 \\
Mobile Video DiT - Mobile~\citep{wu2025taming} & 81.45 & 83.12 & 74.76 \\
Mobile Video DiT - Server~\citep{wu2025taming} & 83.09 & 84.65 & 76.86 \\
CogVideoX 2B~\citep{yang2025cogvideox} & 81.55 & 82.48 & 77.81 \\
PyramidalFlow~\citep{liu2025pyramidalflow} & 81.72 & 84.74 & 69.62 \\
Neodragon~\citep{karnewar2025neodragon} & 81.61 & 83.68 & 73.36 \\
Wan2.1 1.3B~\citep{wan2025} & 83.31 & \underline{85.23} & 75.65 \\
Wan2.1 1.3B*~\citep{wan2025} & 83.10 & 85.10 & 75.12 \\
\cmidrule(l{1.6em}r{0em}){1-4} 
\multicolumn{4}{@{}l}{\hspace{1.6em}\textbf{Linear/Hybrid Models}}\\
\cmidrule(l{1.6em}r{0em}){1-4} 
Efficient VDiT ~\cite{evdit} & 76.14 & - & - \\ 
M4V ~\citep{huang2025m4v} & 81.91 & 83.36 & 76.10 \\
STA~\citep{zhang2025fast} & 83.00 & \textbf{85.37} & 73.52 \\
VSA~\citep{zhang2025faster} & 82.77 & 83.60 & 79.47 \\
SANA-Video~\citep{chen2025sana} & \underline{83.71} & 84.35 & \textbf{81.35} \\
Attention Surgery (15$\times$R2)~\cite{ghafoorian2025attention} & 83.21 & 85.19 & 75.25 \\
\rowcolor{LightCyan}
Wan2.1 1.3B* + ReHyAt (15$\times T_c$=3,$T_o$=1) & \textbf{83.79} & 84.57 & \underline{80.70} \\
\bottomrule
\end{tabular}%
}
\caption{Comparisons with SOTA efficient video diffusion models. `Wan2.1*' is our best reproduction using our evaluation pipeline.}
\label{tab:sota}
\end{table}
\vspace{-6pt}
\section{Results}
\subsection{Generation Quality}
\textbf{VBench SOTA}. Table~\ref{tab:sota} compares ReHyAt model distilled from Wan2.1 1.3B against the state-of-the-art efficient video diffusion models up to 5B parameters. We observe that our method performs very competitively, while forming a chunk-wise RNN that enables running it on mobile. Note that the compute burden to obtain our model is $\sim$160 H100 GPU hours, i.e. less than 1\% of SANA-Video (12 days of 64 H100) and less than 0.01\% of MovieGen~\citep{polyak2024movie}.

\begin{table}[t]
\centering
\footnotesize
\setlength{\tabcolsep}{2pt} 
\resizebox{\columnwidth}{!}{
\begin{tabular}{lcccccc}
\toprule
\multirow{2}{*}{Model} & \multicolumn{6}{c}{VBench-2.0} \\ 
\cmidrule(r){2-7}
 & Total$\uparrow$ & Hum.Fid.$\uparrow$ & Creativity$\uparrow$ & Control.$\uparrow$ & Com.sense$\uparrow$ & Physics$\uparrow$ \\
\midrule
Wan2.1 1.3B             & 56.0 & \underline{80.7} & 48.7 & \textbf{34.0} & \underline{63.4} & \textbf{53.8} \\
CogVideoX-1.5 5B        & 53.4 & 72.1 & 43.7 & 29.6 &  63.2 & 48.2 \\
Attn. Surgery 15$\times$R2 & 55.1 & 78.9 & 47.5 & \underline{33.4} & 63.1 & \underline{52.8} \\
\rowcolor{LightCyan}
ReHyAt 15$\times T_c$=3 & \underline{56.1} & \textbf{81.9} & \underline{55.1} & 30.8 & 62.7 & 50.0 \\
\rowcolor{LightCyan}
ReHyAt 15$\times T_c$=5 & \textbf{56.3} & 79.8 & \textbf{55.7} & 31.9 & \textbf{64.2} & 49.7 \\
\bottomrule
\end{tabular}%
}
\caption{Quantitative comparison on VBench-2.0 benchmark}
\label{tab:vbench2}
\end{table}

\noindent\textbf{VBench2.0}. 
Table~\ref{tab:vbench2} presents the evaluation and comparison of SOTA methods on VBench-2.0 benchmark. While we observe a small drop, ReHyAt still remain competitive to larger models such as CogVideoX1.5 5B.

\begin{table}
\centering
\footnotesize
\begin{tabular}
{@{\extracolsep{\fill}}@{}lccc@{}}
\toprule
\multirow{2}{*}{Prompt Dimension} & \multicolumn{3}{c}{Human Preference \%} \\ \cmidrule(r){2-4}
 & Ours   & No preference & Wan2.1 \\
\midrule
Color & 43.3 & 46.7 & 10.0 \\
Human Action & 21.7 & 41.7 & 36.7 \\
Object Class & 25.0 & 45.0 & 30.0 \\
Overall Consistency & 27.1 & 47.1 & 25.9 \\
Scene & 40.0 & 60.0 & 0.0 \\
Spatial Relationship & 20.0 & 70.0 & 10.0 \\
Subject Consistency & 21.7 & 28.3 & 50.0 \\
Temporal Flickering & 24.0 & 54.0 & 22.0 \\
Temporal Style & 43.3 & 30.0 & 26.7 \\
\midrule
Total & 27.6 & 43.5 & 29.0 \\
\bottomrule
\end{tabular}
\captionof{table}{Results of the method-blinded human visual preference study over 500 paired video comparisons. Rows correspond to subsets filtered by different VBench prompt dimensions.}
\label{tab:human_pref}
\end{table}

\noindent\textbf{Human Preference Evaluation}. Table~\ref{tab:human_pref} shows the results for the human visual preference study comparing our 15$\times T_c$=3 model against the original Wan2.1 1.3B, from a total of 500 paired video comparisons. As can be observed, there is no significant difference between our recurrent hybrid model and the original Wan2.1 in human preference.

\begin{figure*}[t]
    \centering
    \setlength{\tabcolsep}{0pt} 
    \renewcommand{\arraystretch}{0} 
    \begin{tabular}{c}
        \includegraphics[width=0.7\textwidth]{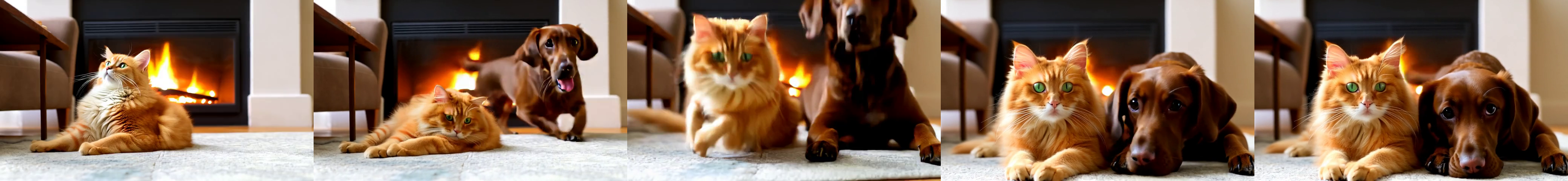} \\
        \includegraphics[width=0.7\textwidth]{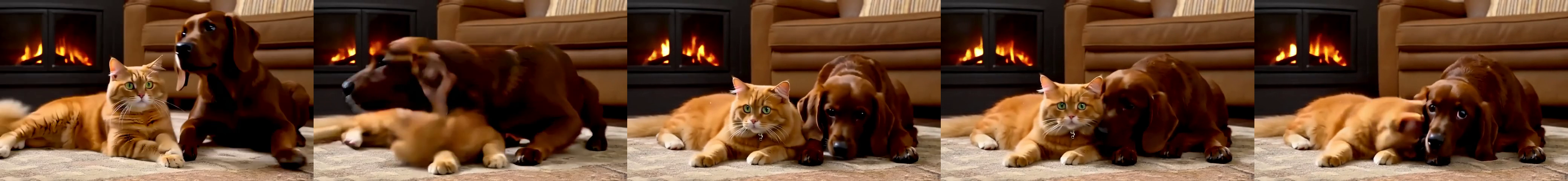} \\
        \includegraphics[width=0.7\textwidth]{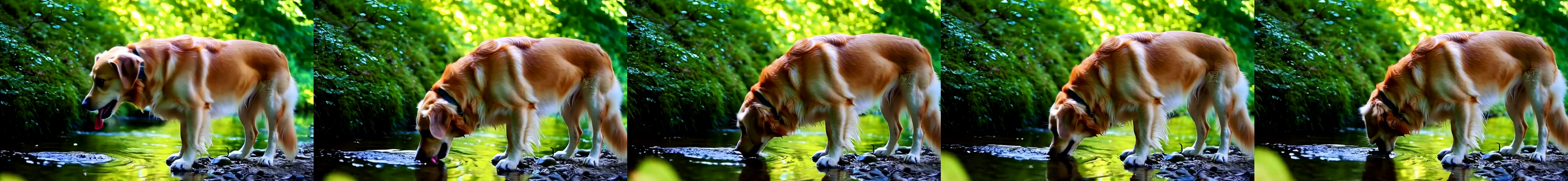} \\
        \includegraphics[width=0.7\textwidth]{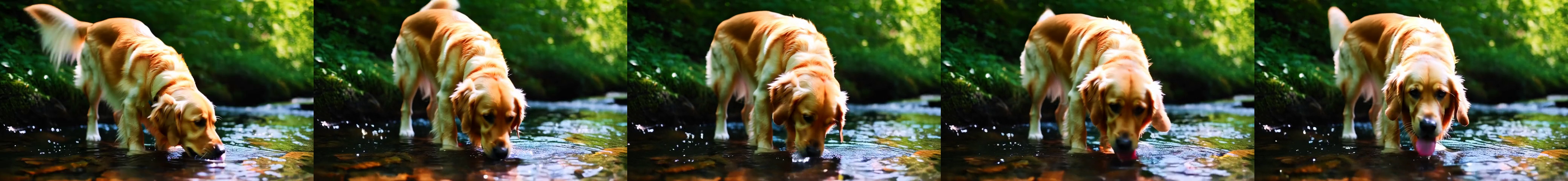}
    \end{tabular}
    \caption{Qualitative comparison of Wan2.1 1.3B (Top) to ReHyAt 15$\times T_c$=3 (bottom) for two sample VBench prompts, ``A cat and a dog.'' and ``A dog drinking water.''} 
    \label{fig:qual}
\end{figure*}
Figure~\ref{fig:qual} shows two qualitative samples and how  ReHyAt compares to Wan2.1 1.3B. More extensive set of qualitative samples are provided in the supplementary materials.
\begin{figure}[t]
\centering
    \includegraphics[width=0.9\linewidth]{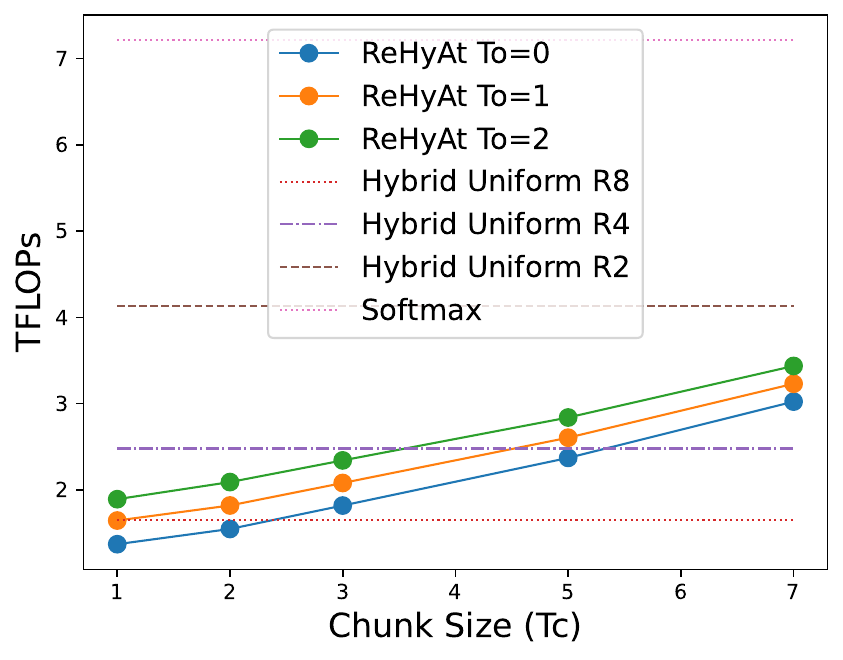}
    \vspace{-10pt}
    \caption{Comparison of attention compute (FLOPs) on 21$\times$30$\times$52 latent size (5 seconds)}
\label{fig:flops}
\end{figure}

\subsection{Sampling Compute Burden}
In Fig.~\ref{fig:flops} we measure how chunk size $T_c$ and chunk overlap size $T_o$ impact the number of floating point operations, also how it compares to flash attention and uniform hybrid attention~\cite{ghafoorian2025attention} with various rates $R \in \{2, 4, 8\}$, as measured on 5s videos at 480$\times$832 resolution, corresponding to a latent size 21$\times$30$\times$52. As can be observed, ReHyAt offers up to 4$\times$ operation saving as compared to flash attention used by Wan2.1. On the other hand, our $T_c$=3, $T_o$=1 model variant remains $\sim$2$\times$ more efficient as compared to the better quality preserving $R=2$ uniform hybrid attention variation.

Fig.~\ref{fig:teaser} top demonstrates how the compute burden grows with increased video duration, comparing the scaling behavior for the Wan2.1 1.3B (flash attention) versus our proposed method (ReHyAt). Here we see that compared to flash attention, our recurrent hybrid attention has a significantly better scaling behavior.
\begin{table}[t]
\centering
\footnotesize
\begin{tabular}{@{}lccccc@{}}
\toprule
& \multicolumn{5}{l}{Number of frames (320$\times$480) resolution}\\ 
\cmidrule(l){2-6}

\multirow{-2}{*}{Attention Block} & \multicolumn{1}{l}{81} & \multicolumn{1}{l}{101} & \multicolumn{1}{l}{121}                            & \multicolumn{1}{l}{141} & \multicolumn{1}{l}{161}\\ 
\midrule

Softmax Flash Attention & 281 & 2964 & 4809 & \textcolor{gray}{OOM} & \textcolor{gray}{OOM} \\
HedgeHog Linear Attention & 360 & 455 & 469 & 542 & \textcolor{gray}{OOM} \\
Uniform Hybrid - R8 & 464 & 625 & 818 & 1215 & \textcolor{gray}{OOM} \\
ReHyAt - $T_c$=3 (ours) & \textbf{192} & \textbf{247} & \textbf{302} & \textbf{329} & \textbf{384} \\ 
\bottomrule
\end{tabular}
\caption{On mobile (Snapdragon8-Gen4) latency (ms) vs. number of frames at 320$\times$480 resolution}
\label{tab:mobile_latency}
\end{table}

Table~\ref{tab:mobile_latency} presents the on-mobile, DiT block latencies in miliseconds for various types of attention mechanism, flash attention, HedgeHog linear attention with learnable $\phi$, uniform hybrid attention $R$=7 and our ReHyAt hybrid method with $T_o$=3, for various video durations from 5s (81 frames) to 10s (161 frames). As can be observed, our recurrent hybrid method is the only one that can easily extend to more than 10s without out-of-memory errors. Within the feasible extent for flash attention (e.g. on 121 frames), our method is $\sim$16$\times$ faster than flash attention used in Wan2.1 1.3B. 

Table~\ref{tab:mobile_memory} shows the memory read/write load that correlates with power consumption and latency. As we observe, due to its more memory-efficient design, our recurrent hybrid attention model is significantly more memory-efficient, e.g. $\sim$11$\times$more efficient in total memory read/write than flash attention at 121 frames ($\sim$7.5s duration). Please note that while the total memory/read write is expected to grow linearly with video duration for ReHyAt, the peak-memory usage remains constant.
\begin{table*}[t]
\centering
\footnotesize
\begin{tabular}{@{}lcccccccccc@{}}
\toprule
& \multicolumn{10}{c}{Number of frames - Memory Read/Write (GB)}\\
\cmidrule(r){2-11}
& \multicolumn{2}{c}{81} & \multicolumn{2}{c}{101} & \multicolumn{2}{c}{121} & \multicolumn{2}{c}{141} & \multicolumn{2}{c}{161} \\
\cmidrule(r){2-11}
\multirow{-3}{*}{Attention Block} & W & R & W & R & W & R & W & R & W & R \\
\midrule
Softmax Flash Attention & 5.1 & 6.0 & 12.9 & 16.4 & \multicolumn{1}{c}{22.7} & 53.6 & \color{gray}{OOM} & \color{gray}{OOM}  & \color{gray}{OOM} & \color{gray}{OOM}\\
HedgeHog Linear Attention & 5.7 & 8.1 & 7.0 & 10.1 & \multicolumn{1}{c}{6.9}  & 11.3 & \multicolumn{1}{c}{8.0} & 13.2 & \color{gray}{OOM} & \color{gray}{OOM} \\
Uniform Hybrid - R8 & 6.3 & 10.1 & 5.2 & 10.9 & 6.4 & 13.2 & 7.8 & 35.2 & \color{gray}{OOM} & \color{gray}{OOM}\\
ReHyAt - $T_c$=3 (ours) & \textbf{1.7} & \textbf{2.8} & \textbf{2.2} & \textbf{3.6} & \textbf{2.7} & \textbf{4.4}  & \multicolumn{1}{c}{\textbf{3.0}} & \textbf{4.8} & \textbf{3.5} & \textbf{5.6}\\
\bottomrule
\end{tabular}
\caption{Comparison of total memory read/write for Wan2.1 DiT Blocks with various attention mechanisms on Snapdragon8-Gen4}
\label{tab:mobile_memory}
\end{table*}

\begin{figure}[t]
    \centering
    \begin{subfigure}[b]{0.40\textwidth}
        \centering
        \includegraphics[width=\textwidth]{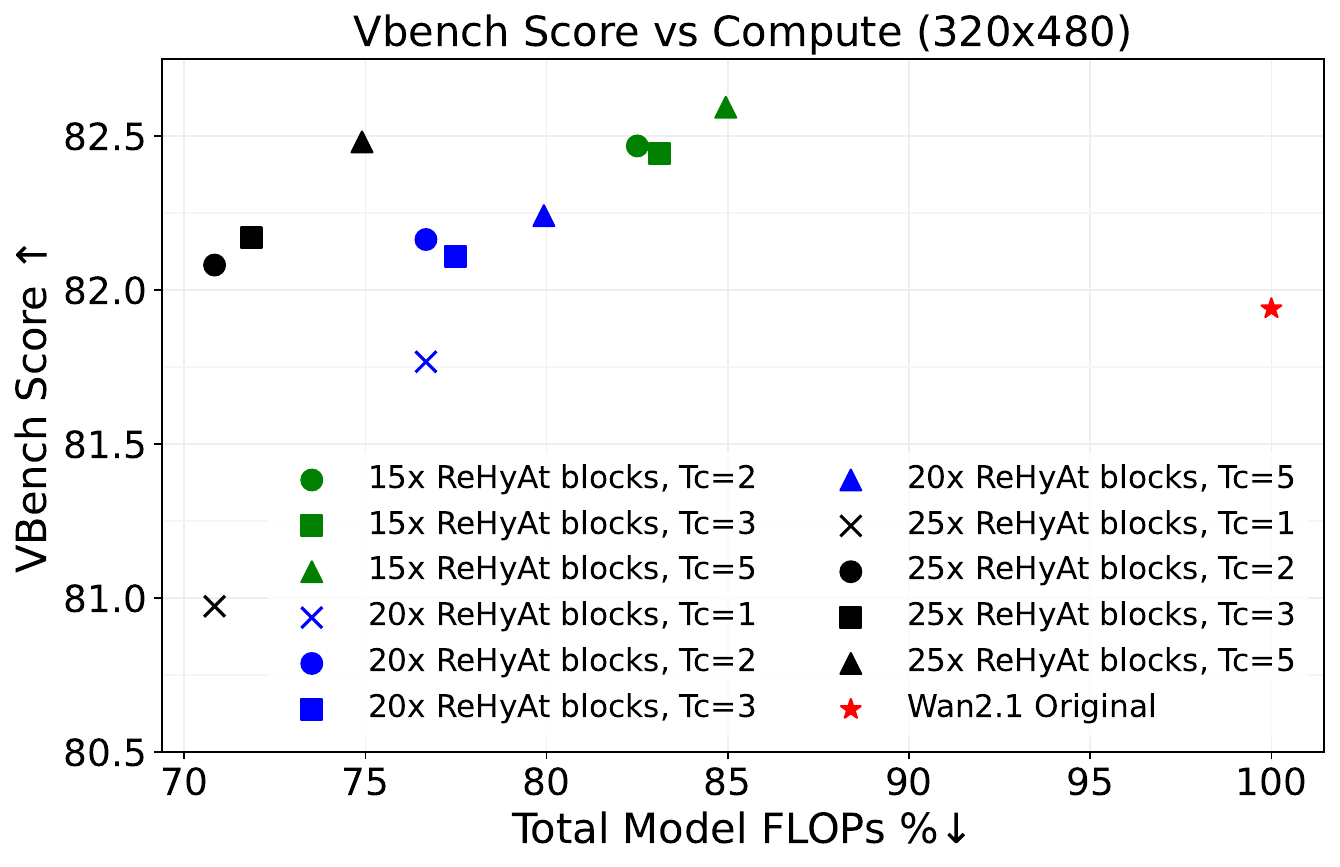}
        \label{fig:320p}
    \end{subfigure}
    \begin{subfigure}[b]{0.40\textwidth}
        \centering
        \includegraphics[width=\textwidth]{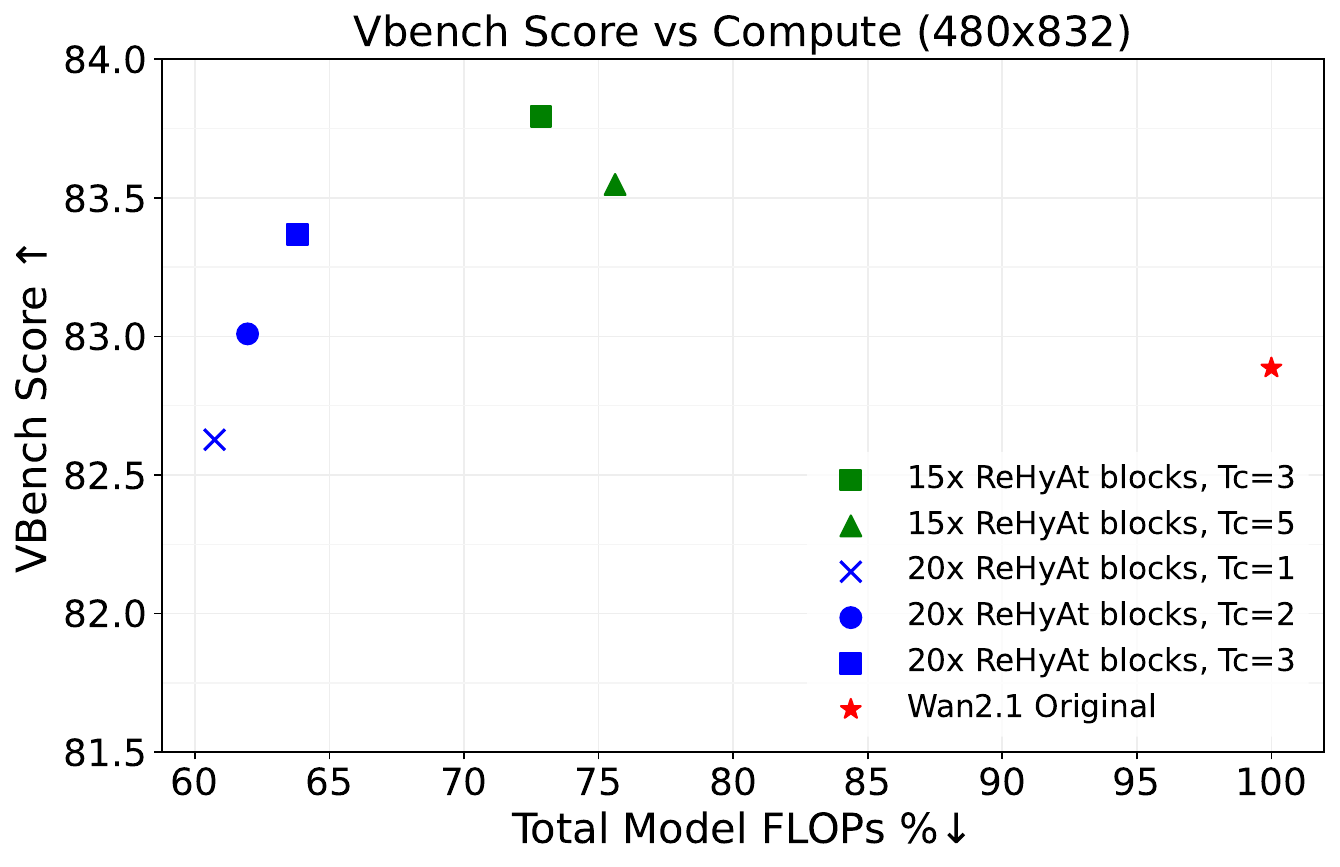}
        \label{fig:480p}
    \end{subfigure}    
    \vspace{-10pt}
    \caption{The total DiT FLOPs percentages versus the VBench score of original Wan2.1 1.3B model compared to various hybrid configurations or 320$\times$480 (top) and 480$\times$832 (bottom) resolutions.}
    \label{fig:vbench_vs_compute}
\end{figure}
\begin{table}[]
\centering
\footnotesize
\begin{tabular}{@{}ccccc@{}}
\toprule

\multirow{2}{*}{\makecell{Chunk-size\\ $T_c$}} & \multirow{2}{*}{\makecell{Block\\ TFLOPs$\downarrow$}} & \multicolumn{3}{c}{VBench} \\ \cmidrule(l){3-5} 
                       & & Total $\uparrow$ & Quality $\uparrow$ & Semantic $\uparrow$ \\ \midrule
1                      & 3.87& 80.97 & 82.37   & 75.39    \\
2                      & 4.04 & 82.08 & 83.86   & 74.99    \\
3                      & 4.30 & 82.17 & 83.72   & \textbf{75.96}    \\
5                      & 4.82 & \textbf{82.48} & \textbf{84.12}   & 75.93    \\ \bottomrule
\end{tabular}
\caption{Impact of $T_c$ on ReHyAt hybrid model quality. All the models have 25$\times$ converted ReHyAt blocks with $T_o$=1.}
\label{tab:ablation_tc}
\end{table}
\subsection{Ablations Studies}
\textbf{Number of ReHyAt Blocks and Chunk-size $T_c$.}
Fig. \ref{fig:vbench_vs_compute} shows scatter plots comparing the computational cost of different variations of ReHyAt, with various number of converted blocks and chunk-sizes $T_c$ as well as the original Wan2.1 1.3B model, at both 320$\times$480 and 480$\times$832 resolutions.
Table~\ref{tab:ablation_tc} shows the Vbench full set evaluation for various $T_c$'s. As expected, increasing $T_c$ generally improves model quality; however, the increase from 1 to 2 yields a more significant improvement compared to further increases to 3 and 4. This is perhaps due to the first extension of the softmax from spatial to spatiotemporal.

\begin{table}[]
\centering
\footnotesize
\begin{tabular}{@{}ccccc@{}}
\toprule

\multirow{2}{*}{\makecell{Chunk-\\overlap $T_o$}} & \multicolumn{4}{c}{VBench} \\ \cmidrule(l){2-5} 
                       & Total $\uparrow$ & Quality $\uparrow$ & Semantic $\uparrow$ & Subj. Cons.$\uparrow$ \\ \midrule
0                      & 81.56 & 83.23 & 74.90 & 90.90 \\
1                      & 82.17 & 83.72 & \textbf{75.96} & 92.05 \\
2                      & 82.17 & 83.84 & 75.50 & 92.13 \\
3                      & \textbf{82.19} & \textbf{83.86} &  75.51 & \textbf{92.24}  \\ \bottomrule
\end{tabular}
\caption{Impact of $T_o$ on ReHyAt hybrid model quality as measured on VBench. All the models have 25$\times$ converted ReHyAt blocks with $T_c$=3.}
\label{tab:ablation_to}
\end{table}
\noindent \textbf{Overlap size $T_o$}. Table~\ref{tab:ablation_to} demonstrates how different chunk overlap size $T_o$ values (ranging from 0 to 3) impacts the generation quality. As anticipated, enabling overlap (i.e., going from $T_o$ = 0 to $T_o$ = 1) results in a notable jump in model quality; however, the total score appears to saturate after that. The mild gradual improvement is still noticeable in the subject consistency dimension. This underlies the importance of overlap mechanism in decreasing temporal incoherencies.

\begin{table}[]
\centering
\footnotesize
\begin{tabular}{@{}ccccc@{}}
\toprule

\multirow{2}{*}{Causal} & \multirow{2}{*}{\makecell{Block\\ TFLOPs$\downarrow$}} & \multicolumn{3}{c}{VBench} \\ \cmidrule(l){3-5} 
                       & & Total $\uparrow$ & Quality $\uparrow$ & Semantic $\uparrow$ \\ \midrule

 $\times$  & 4.17 & 82.27 & 83.84 & \textbf{75.99} \\
 \checkmark & \textbf{4.04} & \textbf{82.35} & \textbf{83.97} & 75.87 \\
\bottomrule
\end{tabular}
\caption{Impact of causality on ReHyAt hybrid model quality as measured on VBench on 15$\times T_c$=3,$T_o$=0 configuration}
\label{tab:ablation_causality}
\end{table}
\noindent \textbf{Causality}.
Table~\ref{tab:ablation_causality} shows the compute and quality metrics for two equal hybrid attention formation, with causality being the only difference. We observe that the additional process to remove the non-causal attention dependency does not deteriorate the quality of the model, at least as measured by VBench. On the other hand, the saving in compute by just removing the forward-looking linear attention is not substantial. The major advantage of causal attention lies in enabling RNN reformulation, in turn enabling lower and constant peak memory and thus on-device generation of longer videos.

\section{Conclusion and Future Work}
In this paper, we introduced ReHyAt, a recurrent hybrid attention mechanism for video diffusion transformers that enables scalable, long-duration video generation with constant memory and linear compute requirements. Our lightweight distillation pipeline achieves near state-of-the-art quality with dramatically reduced training cost. While ReHyAt performs strongly overall, a small fraction of videos—especially with the most efficient variants—still show some temporal incoherence, highlighting an area for future improvement.
{
    \small
    \bibliographystyle{ieeenat_fullname}
    \bibliography{main}

@article{wang2025survey,
title={Survey of Video Diffusion Models: Foundations, Implementations, and Applications},
author={Yimu Wang and Xuye Liu and Wei Pang and Li Ma and Shuai Yuan and Paul Debevec and Ning Yu},
journal={Transactions on Machine Learning Research},
year={2025},
url={https://openreview.net/forum?id=2ODDBObKjH},
}

@article{melnik2024survey,
  title={Video Diffusion Models: A Survey},
  author={Melnik, Andrew and Ljubljanac, Michal and Lu, Cong and Yan, Qi and Ren, Weiming and Ritter, Helge},
  journal={Transactions on Machine Learning Research},
  year={2024},
  url={https://openreview.net/forum?id=rJSHjhEYJx}
}

@inproceedings{peebles2023dit,
  title={Scalable Diffusion Models with Transformers},
  author={Peebles, William and Xie, Saining},
  booktitle={Proceedings of the IEEE/CVF International Conference on Computer Vision},
  year={2023},
  url={https://openaccess.thecvf.com/content/ICCV2023/papers/Peebles_Scalable_Diffusion_Models_with_Transformers_ICCV_2023_paper.pdf}
}

@article{wan2025,
  title={Wan: Open and Advanced Large-Scale Video Generative Models},
  author={Team Wan and others},
  journal={arXiv preprint arXiv:2503.20314},
  year={2025},
  url={https://arxiv.org/abs/2503.20314}
}

@inproceedings{yang2025cogvideox,
title={CogVideoX: Text-to-Video Diffusion Models with An Expert Transformer},
author={Zhuoyi Yang and Jiayan Teng and Wendi Zheng and Ming Ding and Shiyu Huang and Jiazheng Xu and Yuanming Yang and Wenyi Hong and Xiaohan Zhang and Guanyu Feng and Da Yin and Yuxuan,Zhang and Weihan Wang and Yean Cheng and Bin Xu and Xiaotao Gu and Yuxiao Dong and Jie Tang},
booktitle={The Thirteenth International Conference on Learning Representations},
year={2025},
url={https://openreview.net/forum?id=LQzN6TRFg9}
}

@article{tencent2025hunyuan,
  title={HunyuanVideo: A Systematic Framework For Large Video Generation Model},
  author={Tencent AI Lab},
  journal={arXiv preprint arXiv:2412.03603},
  year={2025},
  url={https://arxiv.org/abs/2412.03603}
}

@inproceedings{liu2025pyramidalflow,
title={Pyramidal Flow Matching for Efficient Video Generative Modeling},
author={Yang Jin and Zhicheng Sun and Ningyuan Li and Kun Xu and Kun Xu and Hao Jiang and Nan Zhuang and Quzhe Huang and Yang Song and Yadong MU and Zhouchen Lin},
booktitle={The Thirteenth International Conference on Learning Representations},
year={2025},
url={https://openreview.net/forum?id=66NzcRQuOq}
}

@article{lin2024opensora,
  title={Open-Sora Plan: Open-Source Large Video Generation Model},
  author={Lin, Bin and Ge, Yunyang and Cheng, Xinhua and others},
  journal={arXiv preprint arXiv:2412.00131},
  year={2024},
  url={https://arxiv.org/abs/2412.00131}
}

@inproceedings{vaswani2017attention,
  title={Attention is All You Need},
  author={Vaswani, Ashish and Shazeer, Noam and Parmar, Niki and Uszkoreit, Jakob and Jones, Llion and Gomez, Aidan N and Kaiser, Łukasz and Polosukhin, Illia},
  booktitle={Advances in Neural Information Processing Systems},
  year={2017},
  url={https://arxiv.org/abs/1706.03762}
}

@article{rabe2021memory,
  title={Self-attention Does Not Need $O(n^2)$ Memory},
  author={Rabe, Markus N. and Staats, Charles},
  journal={arXiv preprint arXiv:2112.05682},
  year={2021},
  url={https://arxiv.org/abs/2112.05682}
}

@inproceedings{dao2022flashattention,
  title={FlashAttention: Fast and Memory-Efficient Exact Attention with IO-Awareness},
  author={Dao, Tri and Fu, Daniel Y. and Ermon, Stefano and Rudra, Atri and Re, Christopher},
  booktitle={Advances in Neural Information Processing Systems},
  year={2022},
  url={https://arxiv.org/abs/2205.14135}
}

@inproceedings{katharopoulos2020linear,
  title={Transformers are RNNs: Fast Autoregressive Transformers with Linear Attention},
  author={Katharopoulos, Angelos and Vyas, Apoorv and Pappas, Nikolaos and Fleuret, Fran{\c{c}}ois},
  booktitle={Proceedings of the 37th International Conference on Machine Learning},
  pages={5156--5165},
  year={2020},
  publisher={PMLR},
  url={https://proceedings.mlr.press/v119/katharopoulos20a.html}
}

@inproceedings{choromanski2020performer,
  title={Rethinking Attention with Performers},
  author={Choromanski, Krzysztof and Likhosherstov, Valerii and Dohan, David and Song, Xingyou and Gane, Andreea and Sarlos, Tamas and Hawkins, Peter and Davis, Jared Quincy and Mohiuddin, Afroz and Kaiser, Lukasz and Belanger, David and Colwell, Lucy and Weller, Adrian},
  booktitle={Proceedings of the International Conference on Learning Representations},
  year={2021},
  url={https://arxiv.org/abs/2009.14794}
}

@inproceedings{zhang2024lolcats,
title={Lo{LCAT}s: On Low-Rank Linearizing of Large Language Models},
author={Michael Zhang and Simran Arora and Rahul Chalamala and Benjamin Frederick Spector and Alan Wu and Krithik Ramesh and Aaryan Singhal and Christopher Re},
booktitle={The Thirteenth International Conference on Learning Representations},
year={2025},
url={https://openreview.net/forum?id=8VtGeyJyx9}
}

@inproceedings{zhu2025dig,
  title={DiG: Scalable and Efficient Diffusion Models with Gated Linear Attention},
  author={Zhu, Lianghui and Huang, Zilong and Yan, Hanshu and Feng, Jiashi and Liao, Bencheng and Liew, Jun Hao and Wang, Xinggang},
  booktitle={Proceedings of the IEEE/CVF Conference on Computer Vision and Pattern Recognition},
  year={2025},
  url={https://openaccess.thecvf.com/content/CVPR2025/papers/Zhu_DiG_Scalable_and_Efficient_Diffusion_Models_with_Gated_Linear_Attention_CVPR_2025_paper.pdf}
}

@inproceedings{wang2020linformer,
  title={Linformer: Self-Attention with Linear Complexity},
  author={Wang, Sinong and Li, Belinda and Khabsa, Madian and Fang, Han and Ma, Hao},
  booktitle={Proceedings of the International Conference on Learning Representations},
  year={2020},
  url={https://arxiv.org/abs/2006.04768}
}

@inproceedings{chen2024hedgehog,
title={The Hedgehog \& the Porcupine: Expressive Linear Attentions with Softmax Mimicry},
author={Michael Zhang and Kush Bhatia and Hermann Kumbong and Christopher Re},
booktitle={The Twelfth International Conference on Learning Representations},
year={2024},
url={https://openreview.net/forum?id=4g02l2N2Nx}
}

@inproceedings{li2023sana,
  title={SANA: Efficient Attention for Diffusion Models},
  author={Li, Yifan and others},
  booktitle={Proceedings of the IEEE/CVF International Conference on Computer Vision},
  year={2023},
  url={https://arxiv.org/abs/2303.12345}
}

@inproceedings{yahia2024mobile,
  title={Mobile video diffusion},
  author={Yahia, Haitam Ben and Korzhenkov, Denis and Lelekas, Ioannis and Ghodrati, Amir and Habibian, Amirhossein},
  booktitle={ICCV},
  year={2025}
}

@article{wu2025taming,
  title={Taming Diffusion Transformer for Real-Time Mobile Video Generation},
  author={Wu, Yushu and Li, Yanyu and Kag, Anil and Skorokhodov, Ivan and Menapace, Willi and Ma, Ke and Sahni, Arpit and Hu, Ju and Siarohin, Aliaksandr and Sagar, Dhritiman and others},
  journal={arXiv preprint arXiv:2507.13343},
  year={2025}
}

@article{hacohen2024ltx,
  title={Ltx-video: Realtime video latent diffusion},
  author={HaCohen, Yoav and Chiprut, Nisan and Brazowski, Benny and Shalem, Daniel and Moshe, Dudu and Richardson, Eitan and Levin, Eran and Shiran, Guy and Zabari, Nir and Gordon, Ori and others},
  journal={arXiv preprint arXiv:2501.00103},
  year={2024}
}

@InProceedings{huang2023vbench,
     title={{VBench}: Comprehensive Benchmark Suite for Video Generative Models},
     author={Huang, Ziqi and He, Yinan and Yu, Jiashuo and Zhang, Fan and Si, Chenyang and Jiang, Yuming and Zhang, Yuanhan and Wu, Tianxing and Jin, Qingyang and Chanpaisit, Nattapol and Wang, Yaohui and Chen, Xinyuan and Wang, Limin and Lin, Dahua and Qiao, Yu and Liu, Ziwei},
     booktitle={Proceedings of the IEEE/CVF Conference on Computer Vision and Pattern Recognition},
     year={2024}
 }

@article{zheng2025vbench,
  title={Vbench-2.0: Advancing video generation benchmark suite for intrinsic faithfulness},
  author={Zheng, Dian and Huang, Ziqi and Liu, Hongbo and Zou, Kai and He, Yinan and Zhang, Fan and Zhang, Yuanhan and He, Jingwen and Zheng, Wei-Shi and Qiao, Yu and others},
  journal={arXiv preprint arXiv:2503.21755},
  year={2025}
}

@inproceedings{wu2025snapgen,
  title={Snapgen-v: Generating a five-second video within five seconds on a mobile device},
  author={Wu, Yushu and Zhang, Zhixing and Li, Yanyu and Xu, Yanwu and Kag, Anil and Sui, Yang and Coskun, Huseyin and Ma, Ke and Lebedev, Aleksei and Hu, Ju and others},
  booktitle={Proceedings of the Computer Vision and Pattern Recognition Conference},
  pages={2479--2490},
  year={2025}
}

@article{kim2025device,
  title={On-device Sora: Enabling Training-Free Diffusion-based Text-to-Video Generation for Mobile Devices},
  author={Kim, Bosung and Lee, Kyuhwan and Jeong, Isu and Cheon, Jungmin and Lee, Yeojin and Lee, Seulki},
  journal={arXiv preprint arXiv:2502.04363},
  year={2025}
}

@article{isobe2025amd,
  title={AMD-Hummingbird: Towards an Efficient Text-to-Video Model},
  author={Isobe, Takashi and Cui, He and Zhou, Dong and Ge, Mengmeng and Li, Dong and Barsoum, Emad},
  journal={arXiv preprint arXiv:2503.18559},
  year={2025}
}

@article{evdit,
  title={Efficient-vdit: Efficient video diffusion transformers with attention tile},
  author={Ding, Hangliang and Li, Dacheng and Su, Runlong and Zhang, Peiyuan and Deng, Zhijie and Stoica, Ion and Zhang, Hao},
  journal={arXiv preprint arXiv:2502.06155},
  year={2025}
}

@inproceedings{grafting,
  title={Exploring Diffusion Transformer Designs via Grafting},
  author={Chandrasegaran, Keshigeyan and Poli, Michael and Fu, Daniel Y. and Kim, Dongjun and 
      Hadzic, Lea M. and Li, Manling and Gupta, Agrim and Massaroli, Stefano and 
      Mirhoseini, Azalia and Niebles, Juan Carlos and Ermon, Stefano and Li, Fei-Fei},
  booktitle={NeurIPS},
  year={2025}
}

@inproceedings{cai2022efficientvit,
  title={Efficientvit: Lightweight multi-scale attention for high-resolution dense prediction},
  author={Cai, Han and Li, Junyan and Hu, Muyan and Gan, Chuang and Han, Song},
  booktitle={Proceedings of the IEEE/CVF International Conference on Computer Vision},
  year={2023}
}

@inproceedings{letourneaupadre,
  title={PADRe: A Unifying Polynomial Attention Drop-in Replacement for Efficient Vision Transformer},
  author={Letourneau, Pierre-David and Singh, Manish Kumar and Cheng, Hsin-Pai and Han, Shizhong and Shi, Yunxiao and Jones, Dalton and Langston, Matthew Harper and Cai, Hong and Porikli, Fatih},
  booktitle={The Thirteenth International Conference on Learning Representations},
    year={2025},
}

@inproceedings{wang2025lingen,
  title={Lingen: Towards high-resolution minute-length text-to-video generation with linear computational complexity},
  author={Wang, Hongjie and Ma, Chih-Yao and Liu, Yen-Cheng and Hou, Ji and Xu, Tao and Wang, Jialiang and Juefei-Xu, Felix and Luo, Yaqiao and Zhang, Peizhao and Hou, Tingbo and others},
  booktitle={Proceedings of the Computer Vision and Pattern Recognition Conference},
  pages={2578--2588},
  year={2025}
}

@article{fei2024dimba,
  title={Dimba: Transformer-mamba diffusion models},
  author={Fei, Zhengcong and Fan, Mingyuan and Yu, Changqian and Li, Debang and Zhang, Youqiang and Huang, Junshi},
  journal={arXiv preprint arXiv:2406.01159},
  year={2024}
}

@inproceedings{peruzzo2025adaptor,
  title={ADAPTOR: Adaptive Token Reduction for Video Diffusion Transformers},
  author={Peruzzo, Elia and Karjauv, Adil and Sebe, Nicu and Ghodrati, Amir and Habibian, Amir},
  booktitle={Proceedings of the Computer Vision and Pattern Recognition Conference},
  pages={6365--6371},
  year={2025}
}

@article{karjauv2024movie,
  title={MoViE: Mobile Diffusion for Video Editing},
  author={Karjauv, Adil and Fathima, Noor and Lelekas, Ioannis and Porikli, Fatih and Ghodrati, Amir and Habibian, Amirhossein},
  journal={arXiv preprint arXiv:2412.06578},
  year={2024}
}

@inproceedings{kahatapitiya2024object,
  title={Object-centric diffusion for efficient video editing},
  author={Kahatapitiya, Kumara and Karjauv, Adil and Abati, Davide and Porikli, Fatih and Asano, Yuki M and Habibian, Amirhossein},
  booktitle={European Conference on Computer Vision},
  pages={91--108},
  year={2024},
  organization={Springer}
}

@inproceedings{bolya2023token,
  title={Token merging for fast stable diffusion},
  author={Bolya, Daniel and Hoffman, Judy},
  booktitle={Proceedings of the IEEE/CVF conference on computer vision and pattern recognition},
  pages={4599--4603},
  year={2023}
}

@inproceedings{crowson2024scalable,
  title={Scalable high-resolution pixel-space image synthesis with hourglass diffusion transformers},
  author={Crowson, Katherine and Baumann, Stefan Andreas and Birch, Alex and Abraham, Tanishq Mathew and Kaplan, Daniel Z and Shippole, Enrico},
  booktitle={Forty-first International Conference on Machine Learning},
  year={2024}
}

@article{yang2024parallelizing,
  title={Parallelizing linear transformers with the delta rule over sequence length},
  author={Yang, Songlin and Wang, Bailin and Zhang, Yu and Shen, Yikang and Kim, Yoon},
  journal={Advances in neural information processing systems},
  volume={37},
  pages={115491--115522},
  year={2024}
}

@article{wang2024mamba,
  title={The mamba in the llama: Distilling and accelerating hybrid models},
  author={Wang, Junxiong and Paliotta, Daniele and May, Avner and Rush, Alexander and Dao, Tri},
  journal={Advances in Neural Information Processing Systems},
  volume={37},
  pages={62432--62457},
  year={2024}
}

@inproceedings{mercat2024linearizing,
title={Linearizing Large Language Models},
author={Jean Mercat and Igor Vasiljevic and Sedrick Scott Keh and Kushal Arora and Achal Dave and Adrien Gaidon and Thomas Kollar},
booktitle={First Conference on Language Modeling},
year={2024},
url={https://openreview.net/forum?id=soGxskHGox}
}

@article{fei2024diffusion,
  title={Diffusion-rwkv: Scaling rwkv-like architectures for diffusion models},
  author={Fei, Zhengcong and Fan, Mingyuan and Yu, Changqian and Li, Debang and Huang, Junshi},
  journal={arXiv preprint arXiv:2404.04478},
  year={2024}
}

@article{yao2025diffusion,
  title={Diffusion Transformer-to-Mamba Distillation for High-Resolution Image Generation},
  author={Yao, Yuan and Hong, Yicong and Liu, Difan and Mai, Long and Liu, Feng and Luo, Jiebo},
  journal={arXiv preprint arXiv:2506.18999},
  year={2025}
}

@article{huang2025m4v,
  title={M4V: Multi-Modal Mamba for Text-to-Video Generation},
  author={Huang, Jiancheng and Zhang, Gengwei and Jie, Zequn and Jiao, Siyu and Qian, Yinlong and Chen, Ling and Wei, Yunchao and Ma, Lin},
  journal={arXiv preprint arXiv:2506.10915},
  year={2025}
}

@article{lipman2022flow,
  title={Flow matching for generative modeling},
  author={Lipman, Yaron and Chen, Ricky TQ and Ben-Hamu, Heli and Nickel, Maximilian and Le, Matt},
  journal={arXiv preprint arXiv:2210.02747},
  year={2022}
}

@article{ghafoorian2025attention,
  title={Attention Surgery: An Efficient Recipe to Linearize Your Video Diffusion Transformer},
  author={Ghafoorian, Mohsen and Korzhenkov, Denis and Habibian, Amirhossein},
  journal={arXiv preprint arXiv:2509.24899},
  year={2025}
}

@article{chen2025sana,
  title={SANA-Video: Efficient Video Generation with Block Linear Diffusion Transformer},
  author={Chen, Junsong and Zhao, Yuyang and Yu, Jincheng and Chu, Ruihang and Chen, Junyu and Yang, Shuai and Wang, Xianbang and Pan, Yicheng and Zhou, Daquan and Ling, Huan and others},
  journal={arXiv preprint arXiv:2509.24695},
  year={2025}
}

@article{zhang2024hedgehog,
  title={The hedgehog \& the porcupine: Expressive linear attentions with softmax mimicry},
  author={Zhang, Michael and Bhatia, Kush and Kumbong, Hermann and R{\'e}, Christopher},
  journal={arXiv preprint arXiv:2402.04347},
  year={2024}
}

@article{li2025compact,
  title={Compact Attention: Exploiting Structured Spatio-Temporal Sparsity for Fast Video Generation},
  author={Li, Qirui and Zheng, Guangcong and Zhao, Qi and Li, Jie and Dong, Bin and Yao, Yiwu and Li, Xi},
  journal={arXiv preprint arXiv:2508.12969},
  year={2025}
}

@article{zhang2025fast,
  title={Fast video generation with sliding tile attention},
  author={Zhang, Peiyuan and Chen, Yongqi and Su, Runlong and Ding, Hangliang and Stoica, Ion and Liu, Zhengzhong and Zhang, Hao},
  journal={arXiv preprint arXiv:2502.04507},
  year={2025}
}

@inproceedings{zhang2025faster,
  title={Faster video diffusion with trainable sparse attention},
  author={Zhang, Peiyuan and Chen, Yongqi and Huang, Haofeng and Lin, Will and Liu, Zhengzhong and Stoica, Ion and Xing, Eric P and Zhang, Hao},
  booktitle={The Thirty-ninth Annual Conference on Neural Information Processing Systems},
  year={2025}
}

@article{polyak2024movie,
  title={Movie gen: A cast of media foundation models},
  author={Polyak, Adam and Zohar, Amit and Brown, Andrew and Tjandra, Andros and Sinha, Animesh and Lee, Ann and Vyas, Apoorv and Shi, Bowen and Ma, Chih-Yao and Chuang, Ching-Yao and others},
  journal={arXiv preprint arXiv:2410.13720},
  year={2024}
}

@article{zheng2024open,
  title={Open-sora: Democratizing efficient video production for all},
  author={Zheng, Zangwei and Peng, Xiangyu and Yang, Tianji and Shen, Chenhui and Li, Shenggui and Liu, Hongxin and Zhou, Yukun and Li, Tianyi and You, Yang},
  journal={arXiv preprint arXiv:2412.20404},
  year={2024}
}

@article{karnewar2025neodragon,
    author  = {Animesh Karnewar and Denis Korzhenkov and Ioannis Lelekas and Noor Fathima and Adil Karjauv and Hanwen Xiong, Vancheeswaran Vaidyanathan and Will Zeng and Rafael Esteves and Tushar Singhal and Fatih Porikli and Mohsen Ghafoorian and Amirhossein Habibian},
    title   = {Neodragon: Mobile Video Generation using Diffusion Transformer},
    journal = {arXiv preprint arXiv:2511.06055},
    year    = {2025},
  }
}

\clearpage
\setcounter{page}{1}
\maketitlesupplementary
\section{Appendix}
\subsection{Training Details and Hyperparameters}
Unless stated otherwise in the ablation studies, we parameterize $\phi$ using a two-layer MLP with a polynomial degree of 2. For each hybrid block, we apply separate transformations for keys and queries, denoted as $\phi_k$ and $\phi_q$.

\textbf{Pretraining (Distillation Stage).}  
During pretraining, each block is trained independently while all parameters remain frozen except for $\phi_k$ and $\phi_q$. These are optimized using AdamW with a batch size of 1 and a learning rate of $10^{-3}$, following the value distillation objective described in Equation~(19). Teacher activations for distillation are obtained by sampling with 50 denoising steps and a guidance scale of 5, using the Euler Ancestral Discrete Scheduler to integrate the reverse diffusion process.

\textbf{Finetuning.}  
In the finetuning stage, we update all parameters of the hybrid DiT, including the $\phi$ transformations and feed-forward MLP layers. Training uses AdamW with a batch size of 16, a learning rate of $10^{-5}$, and bf16 mixed-precision. The model is trained for 1{,}000 iterations.

\textbf{Sampling.}  
For generating videos for VBench evaluation, we employ Wan Enhanced prompts and the following sampling configuration: 50 denoising iterations, classifier guidance scale of 6, and the UniPCMultistep noise scheduler with a flow shift of 8.

\subsection{Qualitative Samples}
Figures~\ref{app:qual00}--\ref{app:qual17} present uniformly spaced frames from videos generated by the original Wan2.1 1.3B model and several variants of our recurrent hybrid attention models (15$\times T_c$=5, 15$\times T_c$=3, and 20$\times T_c$=3) across 18 prompts at the original resolution of 480$\times$832. Full video sequences corresponding to these frames are included in the supplementary materials.

\subsection{Detailed VBench Comparison}
\begin{figure*}[t]
\centering
\includegraphics[width=0.7\textwidth]{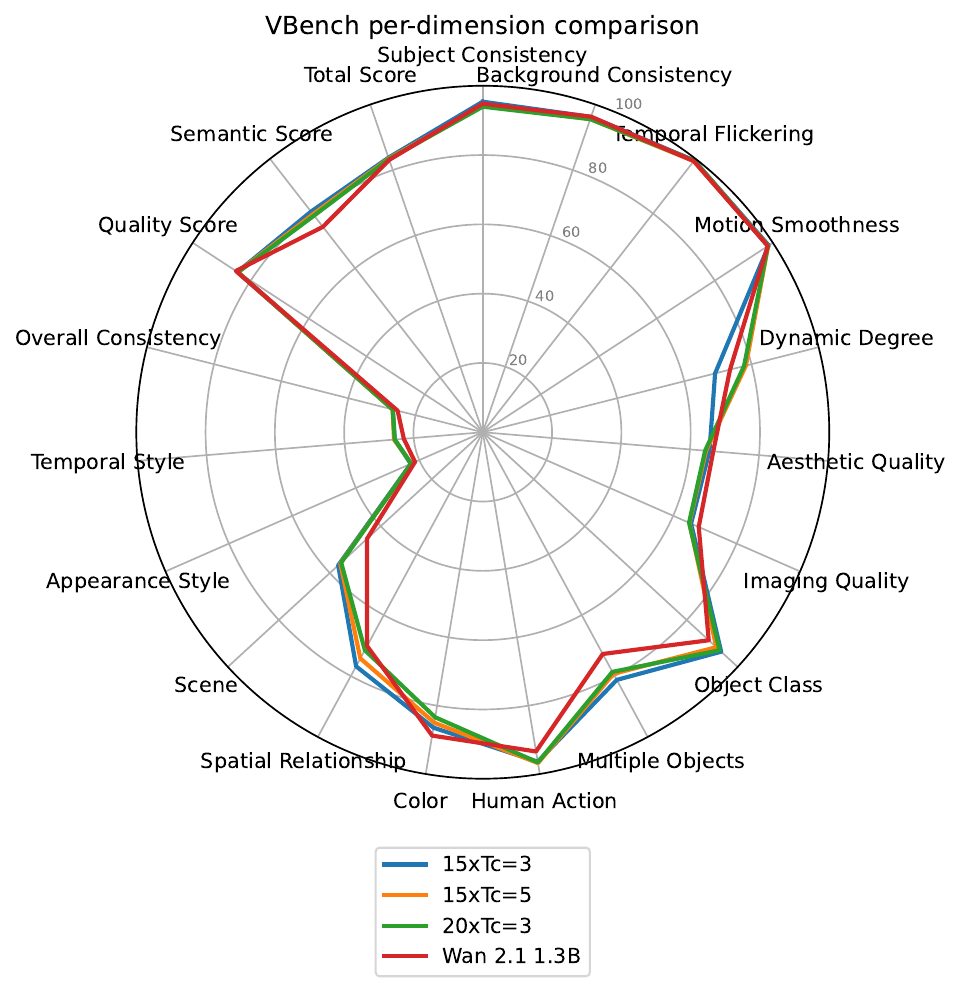}

\caption{Radar plot comparing a subset of our hybrid models with the original Wan 1.3B model on the full VBench set and 480$\times$832 resolution}
\label{app:radar}
\end{figure*}

Figure~\ref{app:radar} compares a selected subset of our hybrid models against Wan2.1 1.3B across all VBench dimensions, evaluated on the full benchmark set at the original resolution (480$\times$832).

\subsection{Detailed VBench-2.0 Comparison}
Tables~\ref{tab:vbench2_part1}--\ref{tab:vbench2_part3} report fine-grained results on the recent VBench-2.0 benchmark at 480$\times$832 resolution. We compare two ReHyAt variants (15$\times T_c$=3 and 15$\times T_c$=5) against Wan2.1 1.3B and attention surgery (15$\times$R2). Both hybrid variants perform on par with Wan2.1 1.3B in terms of the overall Total score.
\begin{table*}[b]
\centering
\resizebox{\textwidth}{!}{%
\begin{tabular}{lcccccccc}
\toprule
\textbf{Method} &
\makecell{\textbf{Human} \\ \textbf{Identity}} &
\makecell{\textbf{Dynamic Spatial} \\ \textbf{Relationship}} &
\makecell{\textbf{Complex} \\ \textbf{Landscape}} &
\makecell{\textbf{Instance} \\ \textbf{Preservation}} &
\makecell{\textbf{Multi-View} \\ \textbf{Consistency}} &
\makecell{\textbf{Human} \\ \textbf{Clothes}} &
\makecell{\textbf{Dynamic} \\ \textbf{Attribute}} &
\makecell{\textbf{Complex} \\ \textbf{Plot}} \\
\midrule
Wan2.1 1.3B$^*$                 & 63.5 & 25.1 & 16.4 & \textbf{86.0} & 9.6 & 97.9 & \textbf{49.1} & 11.3 \\
Attention Surgery (15$\times$R2)& 62.7 & 25.1 & \textbf{18.4} & 84.8 & 7.1 & 97.1 & 44.0 & 13.2 \\
RehHyAt 15$\times T_c$=3        & \textbf{64.7} & \textbf{28.5} & 14.7 & 78.4 & \textbf{12.1} & \textbf{98.1} & 22.0 & 12.7 \\
RehHyAt 15$\times T_c$=5        & 61.6 & 28.0 & 16.7 & 83.6 & 10.6 & 94.2 & 28.6 & \textbf{15.6}  \\
\bottomrule
\end{tabular}
}
\caption{Full VBench-2.0 results (part 1/3).} 
\label{tab:vbench2_part1}
\vspace{0.5em}
\centering
\resizebox{\textwidth}{!}{%
\begin{tabular}{lcccccccc}
\toprule
\textbf{Method} &
\textbf{Mechanics} &
\makecell{\textbf{Human} \\ \textbf{Anatomy}} &
\textbf{Composition} &
\makecell{\textbf{Human} \\ \textbf{Interaction}} &
\makecell{\textbf{Motion} \\ \textbf{Rationality}} &
\textbf{Material} &
\textbf{Diversity} &
\makecell{\textbf{Motion Order} \\ \textbf{Understanding}} \\
\midrule
Wan2.1 1.3B$^*$                 & \textbf{72.4} & 80.6 & 48.4 & 71.7 & 40.8 & 69.4 & 49.1 & 32.0 \\
Attention Surgery (15$\times$R2)& 66.4 & 77.0 & 46.4 & 70.3 & 41.4 & 67.3 & 48.5 & 33.7 \\
RehHyAt 15$\times T_c$=3        & 63.7 & 83.0 & 46.4 & \textbf{75.0} & \textbf{47.1} & \textbf{69.6} & \textbf{63.8} & \textbf{37.0} \\
RehHyAt 15$\times T_c$=5        & 64.7 & \textbf{83.6} & \textbf{51.0} & 72.3 & 44.8 & 67.8 & 60.4 & 34.3 \\
\bottomrule
\end{tabular}
}
\caption{Full VBench-2.0 results (part 2/3).}
\label{tab:vbench2_part2}
\vspace{0.5em}
\centering
\resizebox{\textwidth}{!}{%
\begin{tabular}{lcccccccc}
\toprule
\textbf{Method} &
\makecell{\textbf{Camera} \\ \textbf{Motion}} &
\textbf{Thermotics} &
\makecell{\textbf{Creativity} \\ \textbf{Score}} &
\makecell{\textbf{Commonsense} \\ \textbf{Score}} &
\makecell{\textbf{Controllability} \\ \textbf{Score}} &
\makecell{\textbf{Human Fidelity} \\ \textbf{Score}} &
\makecell{\textbf{Physics} \\ \textbf{Score}} &
\makecell{\textbf{Total} \\ \textbf{Score}} \\
\midrule
Wan2.1 1.3B$^*$ & \textbf{32.1} & 61.7 & 48.7 & 63.4 & \textbf{34.0} & 80.7 & \textbf{53.3} & 56.0 \\
Attention Surgery (15$\times$R2) & 29.0 & \textbf{70.5} & 47.5 & 63.1 & 33.4 & 79.0 & 52.8 & 55.1 \\
RehHyAt 15$\times T_c$=3 & 25.9 & 54.6 & 55.1 & 62.7 & 30.8 & \textbf{81.9} & 50.0 & 56.1 \\
RehHyAt 15$\times T_c$=5 & 29.0 & 55.7 & \textbf{55.7} & \textbf{64.2} & 31.9 & 79.8 & 49.7 & \textbf{56.3} \\
\bottomrule
\end{tabular}
}
\caption{Full VBench-2.0 results (part 3/3).}
\label{tab:vbench2_part3}
\end{table*}

\subsection{Compute complexity vs Attention Surgery}
\begin{figure}[t]
    \centering
    \begin{subfigure}[b]{0.40\textwidth}
        \centering
        \includegraphics[width=\textwidth]{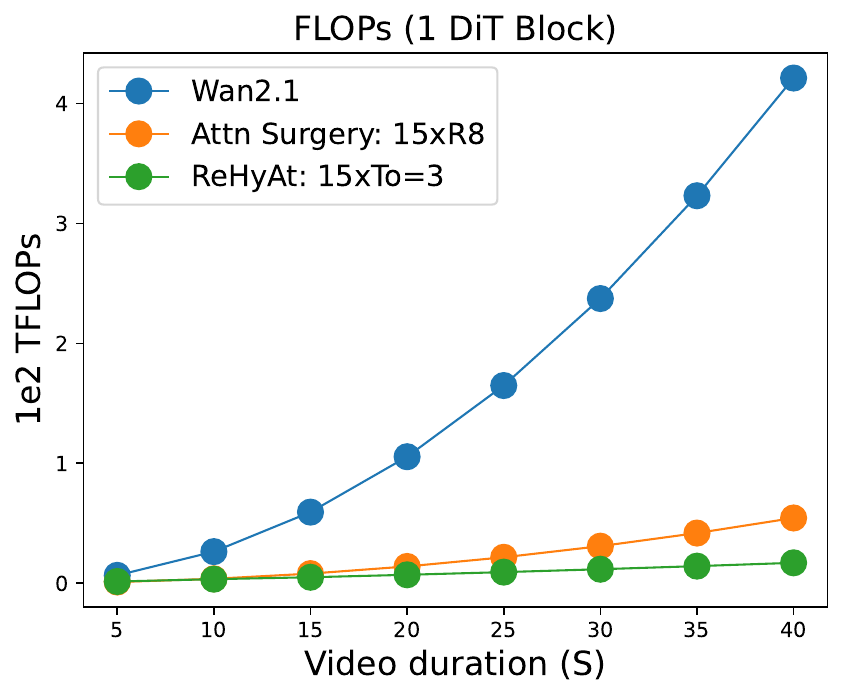}
        \label{fig:320p}
    \end{subfigure}
    \begin{subfigure}[b]{0.40\textwidth}
        \centering
        \includegraphics[width=\textwidth]{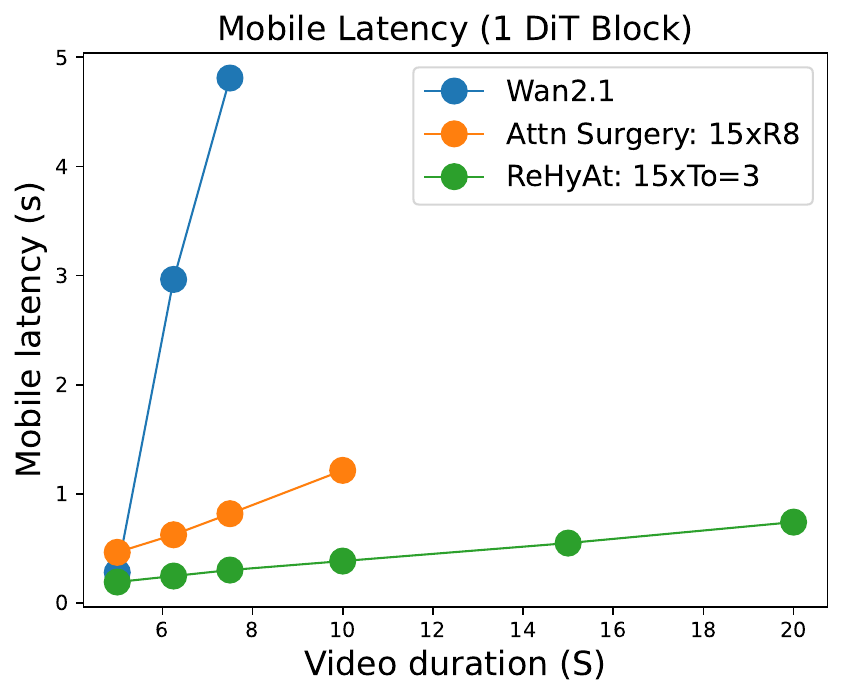}
        \label{fig:480p}
    \end{subfigure}    
    \vspace{-10pt}
    \caption{Compute complexity growth comparisons w.r.t. video length versus Wan2.1 flash attention and attention surgery, in FLOPs (top) and latency (bottom)}
    \label{fig:vs_attn_surgery}
\end{figure}

Figure~\ref{fig:vs_attn_surgery} shows a comparison of our recurrent hybrid attention block in terms of scalability with respect to the video length versus attention surgery hybrid and original Wan2.1 flash attention blocks. 

\subsection{Use of Large Language Models}
We used Microsoft Copilot (a large language model) exclusively to improve clarity and readability. All technical content, experimental design, and conclusions are entirely our own.

\clearpage
\newcommand{\labelgap}{4pt}
\newcommand{\rowsqueeze}{20pt}

\begin{figure*}[htbp]
    \centering
    \setlength{\tabcolsep}{0pt}
    \renewcommand{\arraystretch}{0.1}
    \begin{tabular}{@{}m{0pt}@{}m{\linewidth}@{}}
        \makebox[0pt][r]{\raisebox{0pt}[0pt][0pt]{\rotatebox{90}{\scriptsize\hspace{-20pt}Wan2.1 1.3B}}\hspace{\labelgap}} &
        \includegraphics[width=\linewidth]{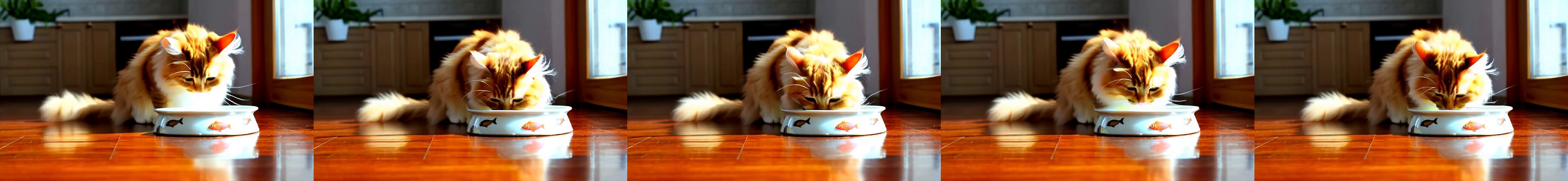} \\
        [\rowsqueeze]

        \makebox[0pt][r]{\raisebox{0pt}[0pt][0pt]{\rotatebox{90}{\scriptsize\hspace{-10pt}15$\times T_c$=5}}\hspace{\labelgap}} &
        \includegraphics[width=\linewidth]{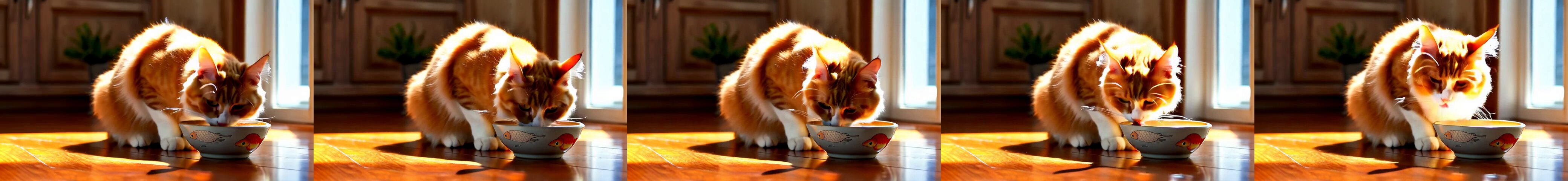} \\
        [\rowsqueeze]

        \makebox[0pt][r]{\raisebox{0pt}[0pt][0pt]{\rotatebox{90}{\scriptsize\hspace{-10pt}15$\times T_c$=3}}\hspace{\labelgap}} &
        \includegraphics[width=\linewidth]{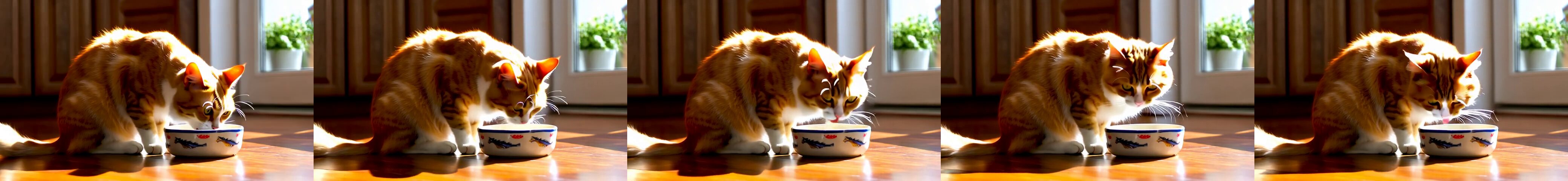} \\
        [\rowsqueeze]

        \makebox[0pt][r]{\raisebox{0pt}[0pt][0pt]{\rotatebox{90}{\scriptsize\hspace{-10pt}20$\times T_c$=3}}\hspace{\labelgap}} &
        \includegraphics[width=\linewidth]{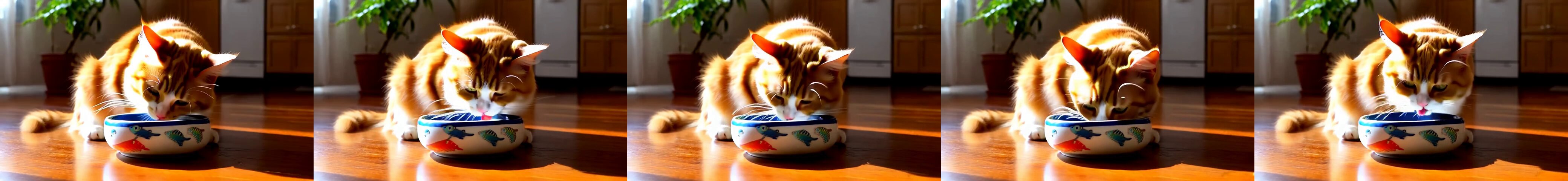} \\
        [\rowsqueeze]

    \end{tabular}
    \caption{Qualitative videos comparing original Wan2.1 1.3B model to our various hybrid variations for input prompt \emph{A cat eating food out of a bowl}}%
    \label{app:qual00}
\end{figure*}

\begin{figure*}[htbp]
    \centering
    \setlength{\tabcolsep}{0pt}
    \renewcommand{\arraystretch}{0.1}
    \begin{tabular}{@{}m{0pt}@{}m{\linewidth}@{}}
        \makebox[0pt][r]{\raisebox{0pt}[0pt][0pt]{\rotatebox{90}{\scriptsize\hspace{-20pt}Wan2.1 1.3B}}\hspace{\labelgap}} &
        \includegraphics[width=\linewidth]{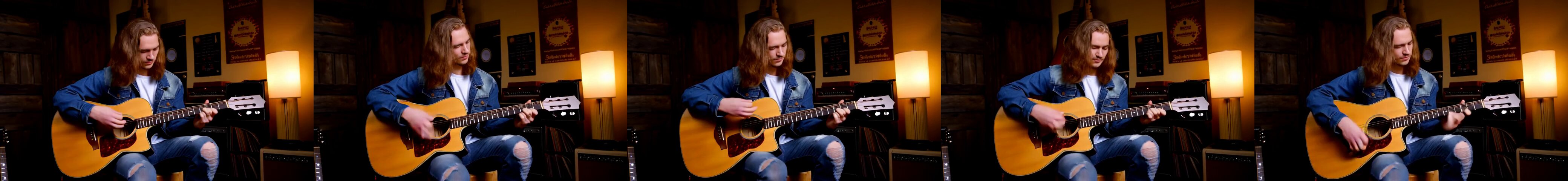} \\
        [\rowsqueeze]

        \makebox[0pt][r]{\raisebox{0pt}[0pt][0pt]{\rotatebox{90}{\scriptsize\hspace{-10pt}15$\times T_c$=5}}\hspace{\labelgap}} &
        \includegraphics[width=\linewidth]{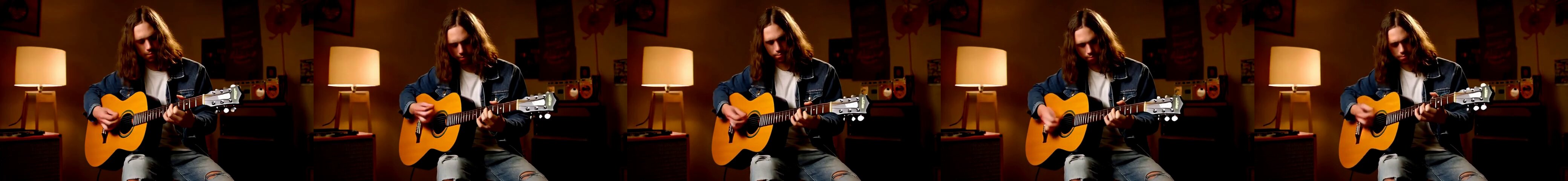} \\
        [\rowsqueeze]

        \makebox[0pt][r]{\raisebox{0pt}[0pt][0pt]{\rotatebox{90}{\scriptsize\hspace{-10pt}15$\times T_c$=3}}\hspace{\labelgap}} &
        \includegraphics[width=\linewidth]{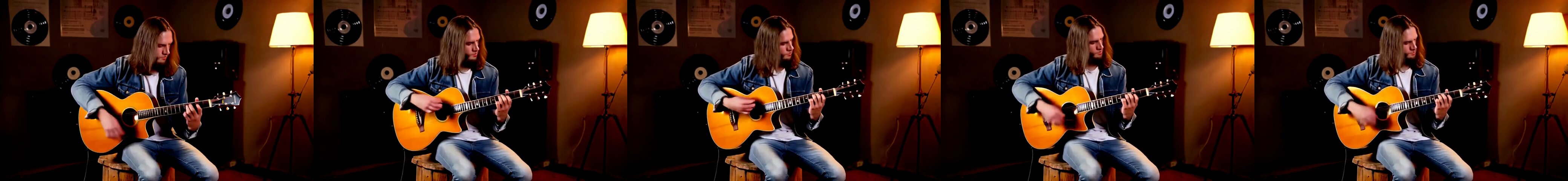} \\
        [\rowsqueeze]

        \makebox[0pt][r]{\raisebox{0pt}[0pt][0pt]{\rotatebox{90}{\scriptsize\hspace{-10pt}20$\times T_c$=3}}\hspace{\labelgap}} &
        \includegraphics[width=\linewidth]{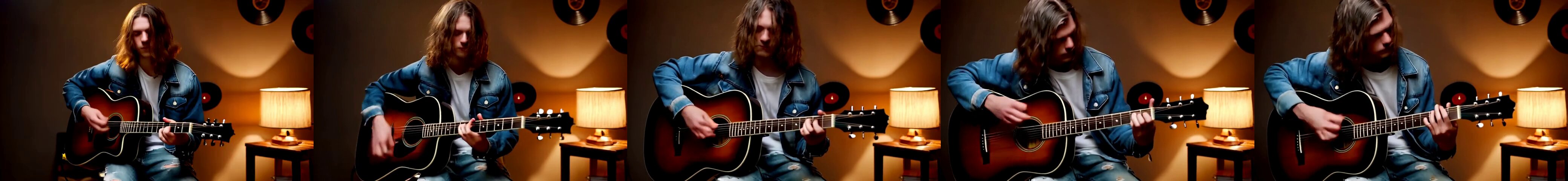} \\
        [\rowsqueeze]

    \end{tabular}
    \caption{Qualitative videos comparing original Wan2.1 1.3B model to our various hybrid variations for input prompt \emph{a person playing guitar}}%
    \label{app:qual01}
\end{figure*}

\begin{figure*}[htbp]
    \centering
    \setlength{\tabcolsep}{0pt}
    \renewcommand{\arraystretch}{0.1}
    \begin{tabular}{@{}m{0pt}@{}m{\linewidth}@{}}
        \makebox[0pt][r]{\raisebox{0pt}[0pt][0pt]{\rotatebox{90}{\scriptsize\hspace{-20pt}Wan2.1 1.3B}}\hspace{\labelgap}} &
        \includegraphics[width=\linewidth]{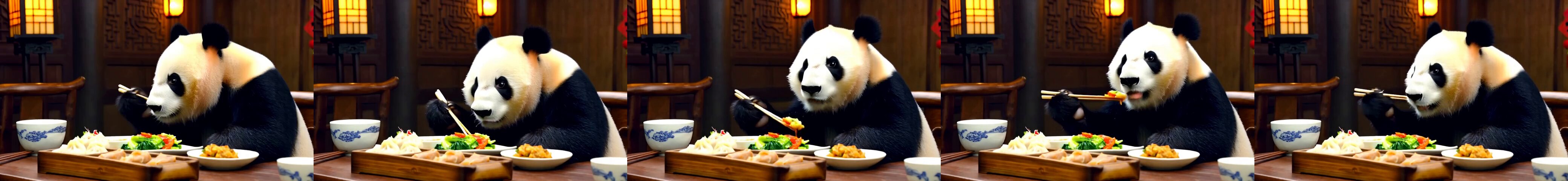} \\
        [\rowsqueeze]

        \makebox[0pt][r]{\raisebox{0pt}[0pt][0pt]{\rotatebox{90}{\scriptsize\hspace{-10pt}15$\times T_c$=5}}\hspace{\labelgap}} &
        \includegraphics[width=\linewidth]{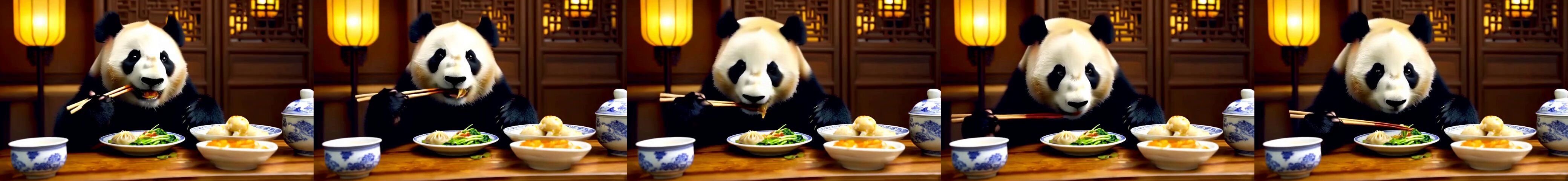} \\
        [\rowsqueeze]

        \makebox[0pt][r]{\raisebox{0pt}[0pt][0pt]{\rotatebox{90}{\scriptsize\hspace{-10pt}15$\times T_c$=3}}\hspace{\labelgap}} &
        \includegraphics[width=\linewidth]{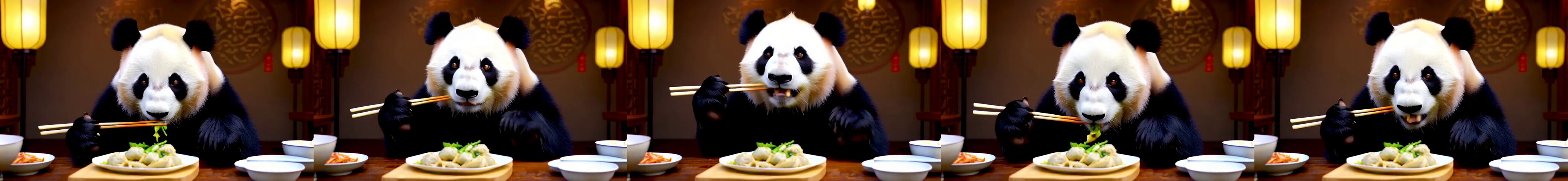} \\
        [\rowsqueeze]

        \makebox[0pt][r]{\raisebox{0pt}[0pt][0pt]{\rotatebox{90}{\scriptsize\hspace{-10pt}20$\times T_c$=3}}\hspace{\labelgap}} &
        \includegraphics[width=\linewidth]{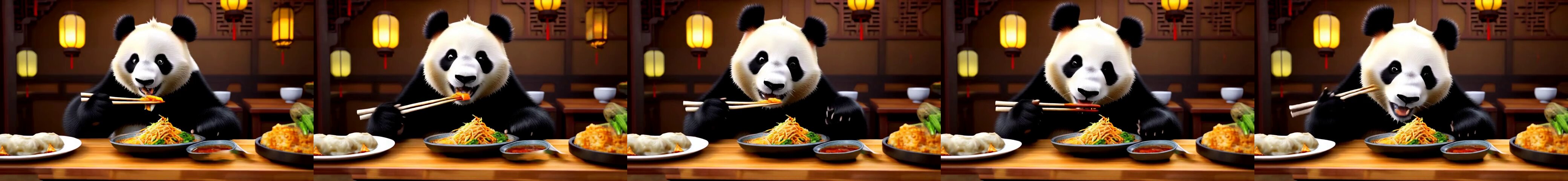} \\
        [\rowsqueeze]

    \end{tabular}
    \caption{Qualitative videos comparing original Wan2.1 1.3B model to our various hybrid variations for input prompt \emph{A cute fluffy panda eating Chinese food in a restaurant}}%
    \label{app:qual02}
\end{figure*}

\begin{figure*}[htbp]
    \centering
    \setlength{\tabcolsep}{0pt}
    \renewcommand{\arraystretch}{0.1}
    \begin{tabular}{@{}m{0pt}@{}m{\linewidth}@{}}
        \makebox[0pt][r]{\raisebox{0pt}[0pt][0pt]{\rotatebox{90}{\scriptsize\hspace{-20pt}Wan2.1 1.3B}}\hspace{\labelgap}} &
        \includegraphics[width=\linewidth]{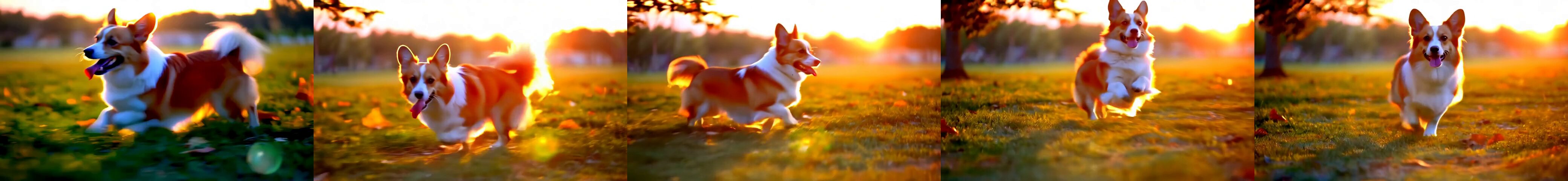} \\
        [\rowsqueeze]

        \makebox[0pt][r]{\raisebox{0pt}[0pt][0pt]{\rotatebox{90}{\scriptsize\hspace{-10pt}15$\times T_c$=5}}\hspace{\labelgap}} &
        \includegraphics[width=\linewidth]{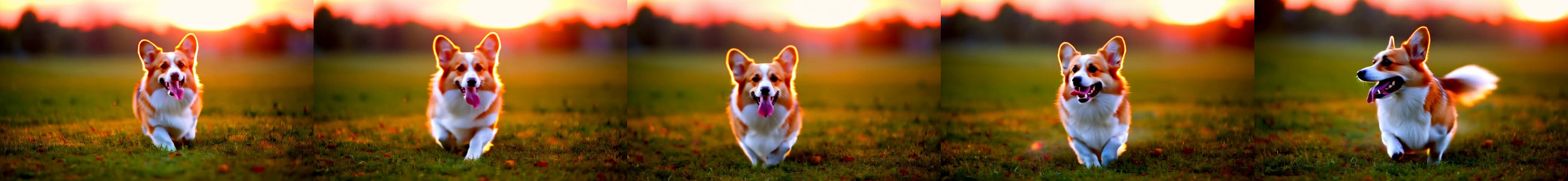} \\
        [\rowsqueeze]

        \makebox[0pt][r]{\raisebox{0pt}[0pt][0pt]{\rotatebox{90}{\scriptsize\hspace{-10pt}15$\times T_c$=3}}\hspace{\labelgap}} &
        \includegraphics[width=\linewidth]{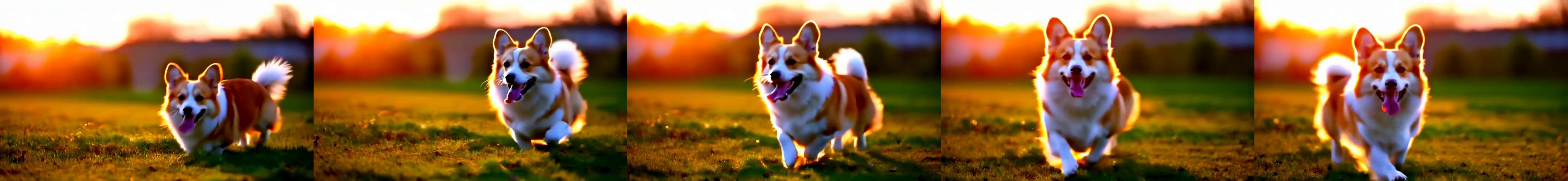} \\
        [\rowsqueeze]

        \makebox[0pt][r]{\raisebox{0pt}[0pt][0pt]{\rotatebox{90}{\scriptsize\hspace{-10pt}20$\times T_c$=3}}\hspace{\labelgap}} &
        \includegraphics[width=\linewidth]{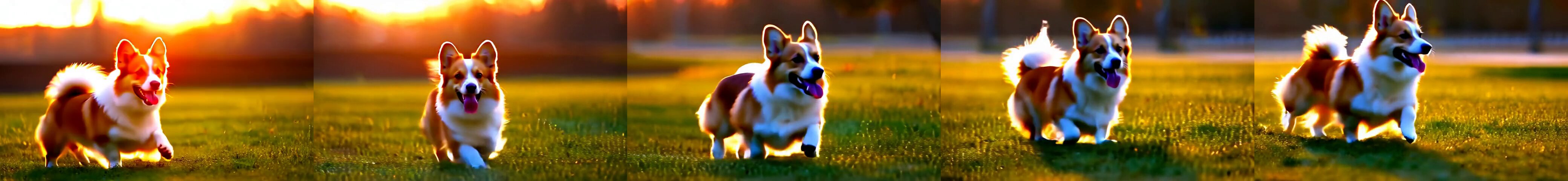} \\
        [\rowsqueeze]

    \end{tabular}
    \caption{Qualitative videos comparing original Wan2.1 1.3B model to our various hybrid variations for input prompt \emph{A cute happy Corgi playing in park, sunset, with an intense shaking effect}}%
    \label{app:qual03}
\end{figure*}

\begin{figure*}[htbp]
    \centering
    \setlength{\tabcolsep}{0pt}
    \renewcommand{\arraystretch}{0.1}
    \begin{tabular}{@{}m{0pt}@{}m{\linewidth}@{}}
        \makebox[0pt][r]{\raisebox{0pt}[0pt][0pt]{\rotatebox{90}{\scriptsize\hspace{-20pt}Wan2.1 1.3B}}\hspace{\labelgap}} &
        \includegraphics[width=\linewidth]{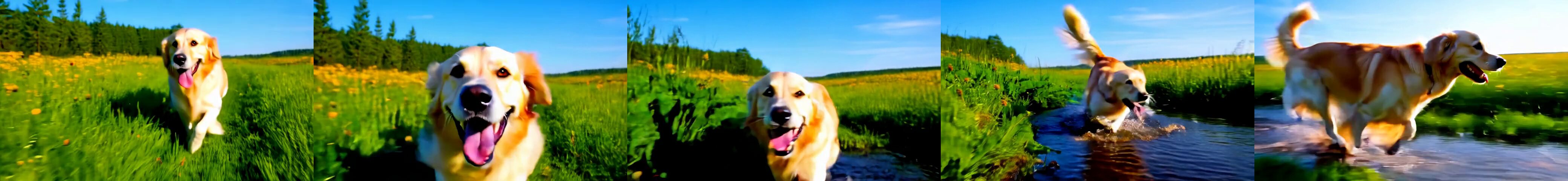} \\
        [\rowsqueeze]

        \makebox[0pt][r]{\raisebox{0pt}[0pt][0pt]{\rotatebox{90}{\scriptsize\hspace{-10pt}15$\times T_c$=5}}\hspace{\labelgap}} &
        \includegraphics[width=\linewidth]{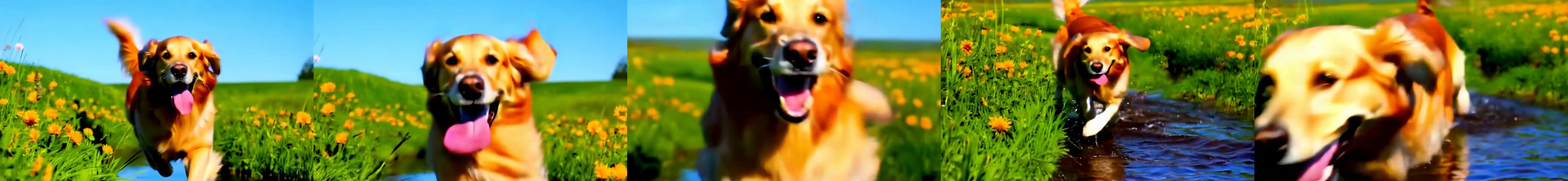} \\
        [\rowsqueeze]

        \makebox[0pt][r]{\raisebox{0pt}[0pt][0pt]{\rotatebox{90}{\scriptsize\hspace{-10pt}15$\times T_c$=3}}\hspace{\labelgap}} &
        \includegraphics[width=\linewidth]{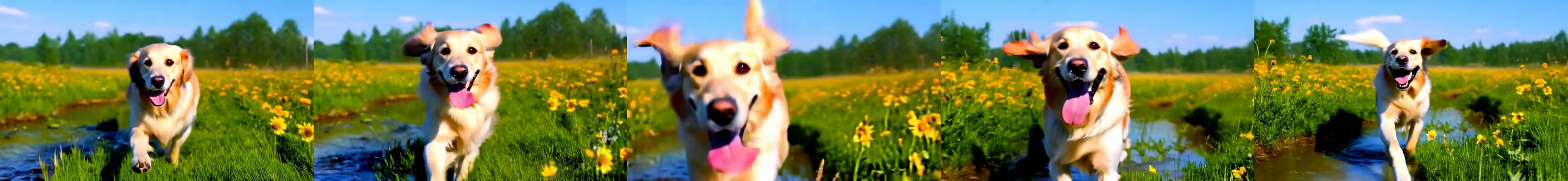} \\
        [\rowsqueeze]

        \makebox[0pt][r]{\raisebox{0pt}[0pt][0pt]{\rotatebox{90}{\scriptsize\hspace{-10pt}20$\times T_c$=3}}\hspace{\labelgap}} &
        \includegraphics[width=\linewidth]{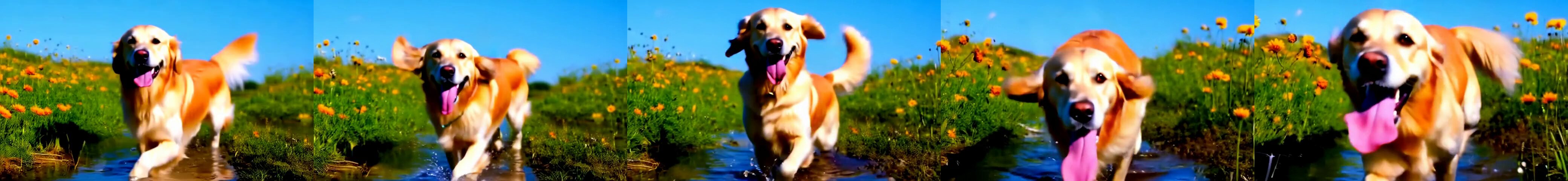} \\
        [\rowsqueeze]

    \end{tabular}
    \caption{Qualitative videos comparing original Wan2.1 1.3B model to our various hybrid variations for input prompt \emph{a dog running happily}}%
    \label{app:qual04}
\end{figure*}

\begin{figure*}[htbp]
    \centering
    \setlength{\tabcolsep}{0pt}
    \renewcommand{\arraystretch}{0.1}
    \begin{tabular}{@{}m{0pt}@{}m{\linewidth}@{}}
        \makebox[0pt][r]{\raisebox{0pt}[0pt][0pt]{\rotatebox{90}{\scriptsize\hspace{-20pt}Wan2.1 1.3B}}\hspace{\labelgap}} &
        \includegraphics[width=\linewidth]{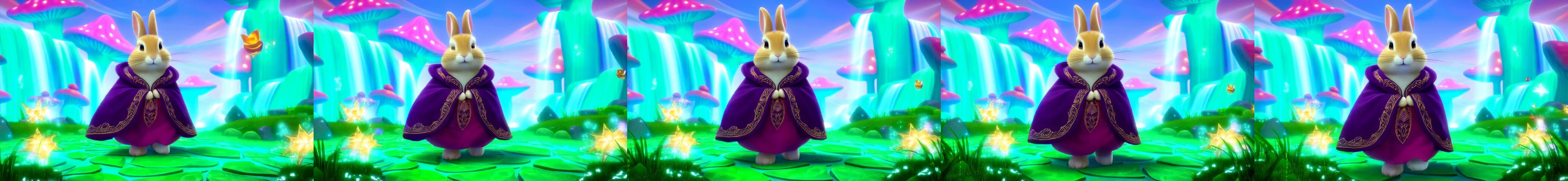} \\
        [\rowsqueeze]

        \makebox[0pt][r]{\raisebox{0pt}[0pt][0pt]{\rotatebox{90}{\scriptsize\hspace{-10pt}15$\times T_c$=5}}\hspace{\labelgap}} &
        \includegraphics[width=\linewidth]{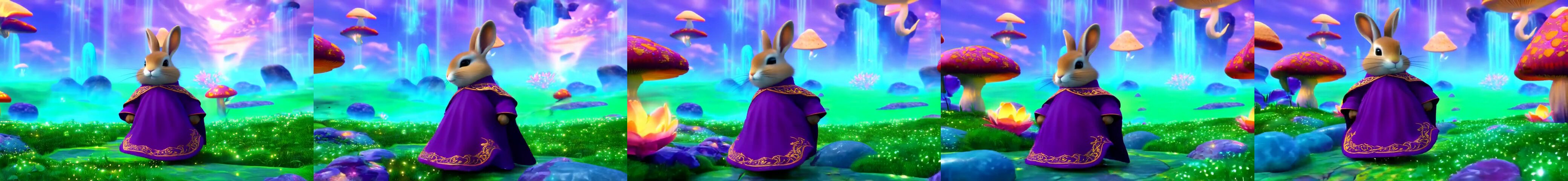} \\
        [\rowsqueeze]

        \makebox[0pt][r]{\raisebox{0pt}[0pt][0pt]{\rotatebox{90}{\scriptsize\hspace{-10pt}15$\times T_c$=3}}\hspace{\labelgap}} &
        \includegraphics[width=\linewidth]{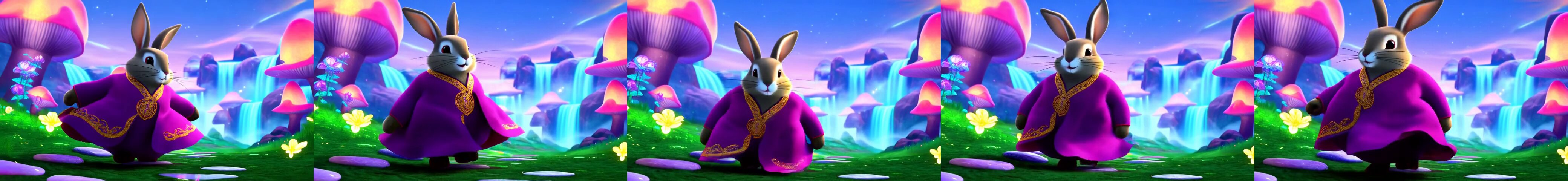} \\
        [\rowsqueeze]

        \makebox[0pt][r]{\raisebox{0pt}[0pt][0pt]{\rotatebox{90}{\scriptsize\hspace{-10pt}20$\times T_c$=3}}\hspace{\labelgap}} &
        \includegraphics[width=\linewidth]{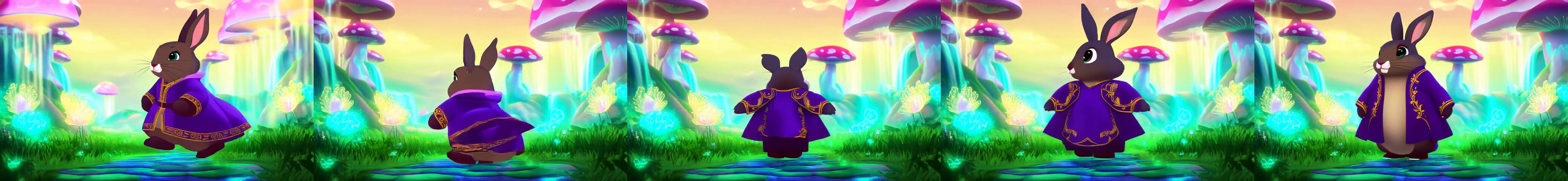} \\
        [\rowsqueeze]

    \end{tabular}
    \caption{Qualitative videos comparing original Wan2.1 1.3B model to our various hybrid variations for input prompt \emph{A fat rabbit wearing a purple robe walking through a fantasy landscape.}}%
    \label{app:qual05}
\end{figure*}

\begin{figure*}[htbp]
    \centering
    \setlength{\tabcolsep}{0pt}
    \renewcommand{\arraystretch}{0.1}
    \begin{tabular}{@{}m{0pt}@{}m{\linewidth}@{}}
        \makebox[0pt][r]{\raisebox{0pt}[0pt][0pt]{\rotatebox{90}{\scriptsize\hspace{-20pt}Wan2.1 1.3B}}\hspace{\labelgap}} &
        \includegraphics[width=\linewidth]{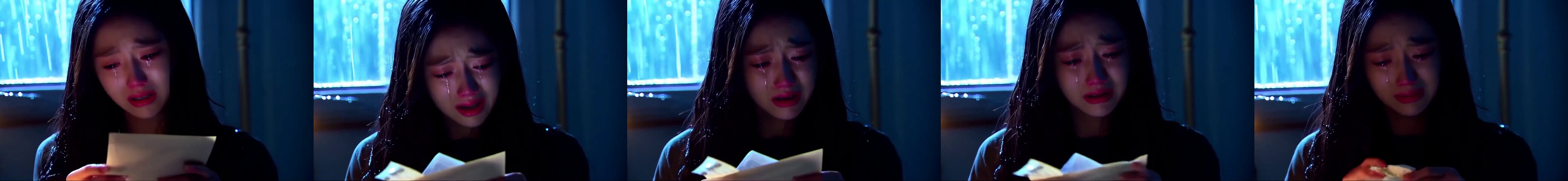} \\
        [\rowsqueeze]

        \makebox[0pt][r]{\raisebox{0pt}[0pt][0pt]{\rotatebox{90}{\scriptsize\hspace{-10pt}15$\times T_c$=5}}\hspace{\labelgap}} &
        \includegraphics[width=\linewidth]{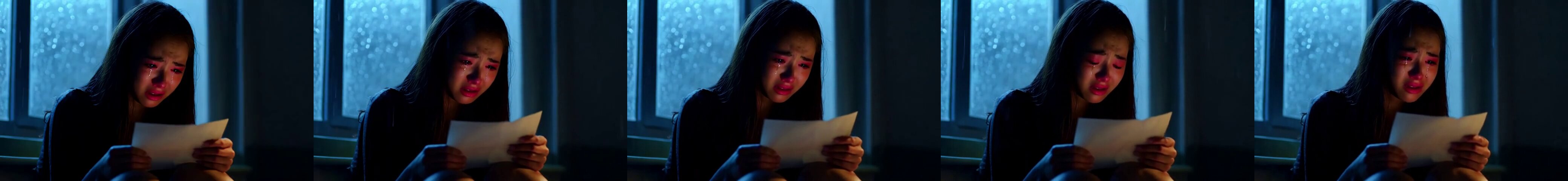} \\
        [\rowsqueeze]

        \makebox[0pt][r]{\raisebox{0pt}[0pt][0pt]{\rotatebox{90}{\scriptsize\hspace{-10pt}15$\times T_c$=3}}\hspace{\labelgap}} &
        \includegraphics[width=\linewidth]{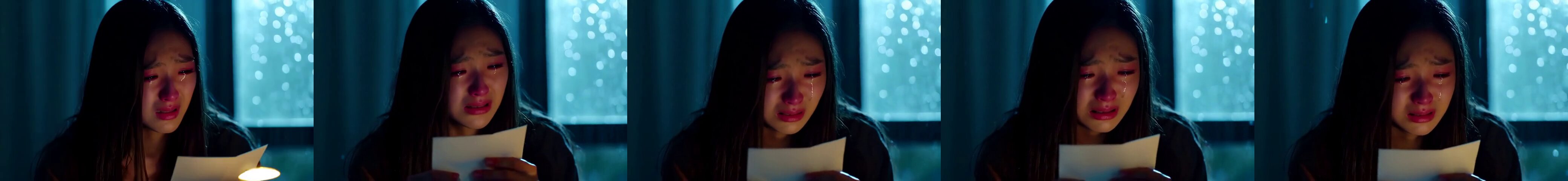} \\
        [\rowsqueeze]

        \makebox[0pt][r]{\raisebox{0pt}[0pt][0pt]{\rotatebox{90}{\scriptsize\hspace{-10pt}20$\times T_c$=3}}\hspace{\labelgap}} &
        \includegraphics[width=\linewidth]{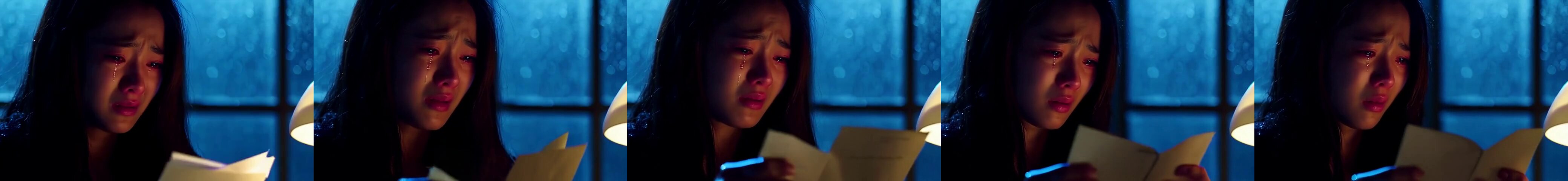} \\
        [\rowsqueeze]

    \end{tabular}
    \caption{Qualitative videos comparing original Wan2.1 1.3B model to our various hybrid variations for input prompt \emph{A person is crying}}%
    \label{app:qual06}
\end{figure*}

\begin{figure*}[htbp]
    \centering
    \setlength{\tabcolsep}{0pt}
    \renewcommand{\arraystretch}{0.1}
    \begin{tabular}{@{}m{0pt}@{}m{\linewidth}@{}}
        \makebox[0pt][r]{\raisebox{0pt}[0pt][0pt]{\rotatebox{90}{\scriptsize\hspace{-20pt}Wan2.1 1.3B}}\hspace{\labelgap}} &
        \includegraphics[width=\linewidth]{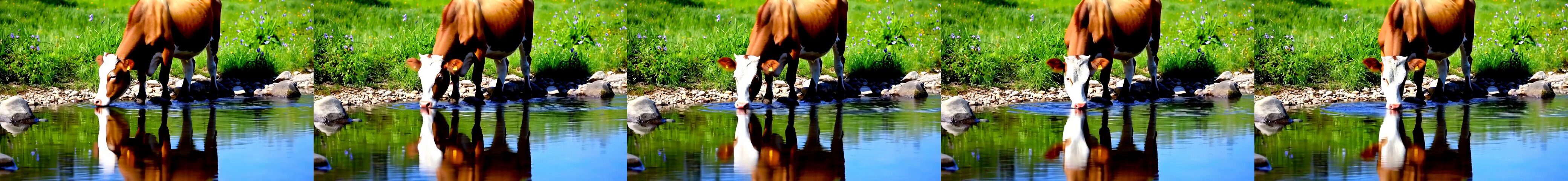} \\
        [\rowsqueeze]

        \makebox[0pt][r]{\raisebox{0pt}[0pt][0pt]{\rotatebox{90}{\scriptsize\hspace{-10pt}15$\times T_c$=5}}\hspace{\labelgap}} &
        \includegraphics[width=\linewidth]{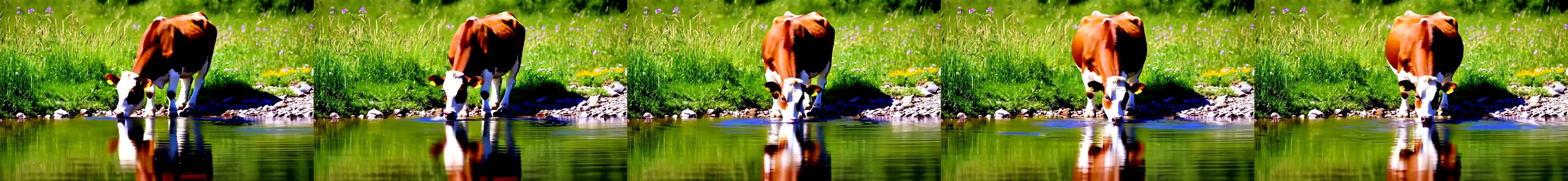} \\
        [\rowsqueeze]

        \makebox[0pt][r]{\raisebox{0pt}[0pt][0pt]{\rotatebox{90}{\scriptsize\hspace{-10pt}15$\times T_c$=3}}\hspace{\labelgap}} &
        \includegraphics[width=\linewidth]{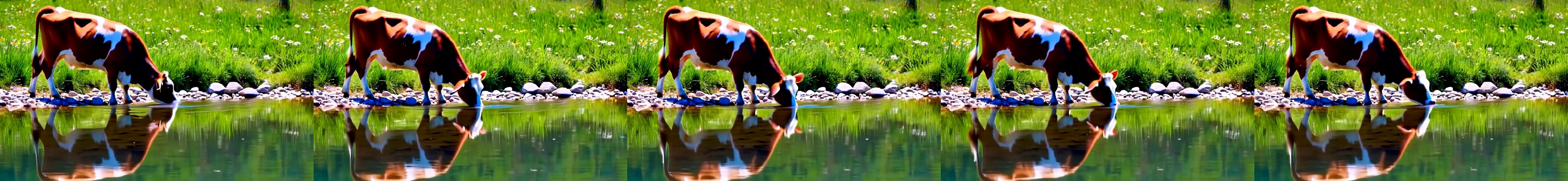} \\
        [\rowsqueeze]

        \makebox[0pt][r]{\raisebox{0pt}[0pt][0pt]{\rotatebox{90}{\scriptsize\hspace{-10pt}20$\times T_c$=3}}\hspace{\labelgap}} &
        \includegraphics[width=\linewidth]{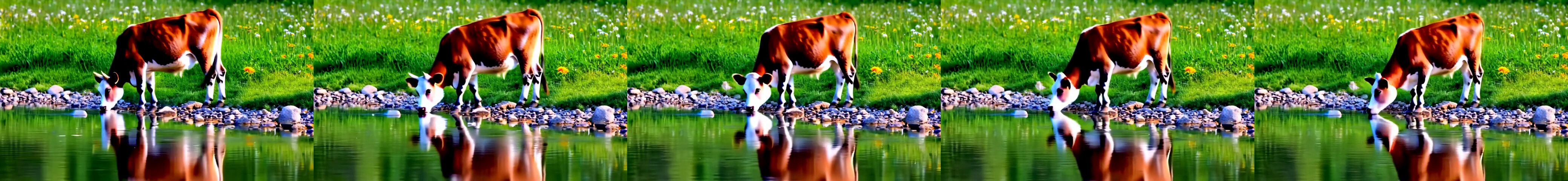} \\
        [\rowsqueeze]

    \end{tabular}
    \caption{Qualitative videos comparing original Wan2.1 1.3B model to our various hybrid variations for input prompt \emph{a cow bending down to drink water from a river}}%
    \label{app:qual08}
\end{figure*}

\begin{figure*}[htbp]
    \centering
    \setlength{\tabcolsep}{0pt}
    \renewcommand{\arraystretch}{0.1}
    \begin{tabular}{@{}m{0pt}@{}m{\linewidth}@{}}
        \makebox[0pt][r]{\raisebox{0pt}[0pt][0pt]{\rotatebox{90}{\scriptsize\hspace{-20pt}Wan2.1 1.3B}}\hspace{\labelgap}} &
        \includegraphics[width=\linewidth]{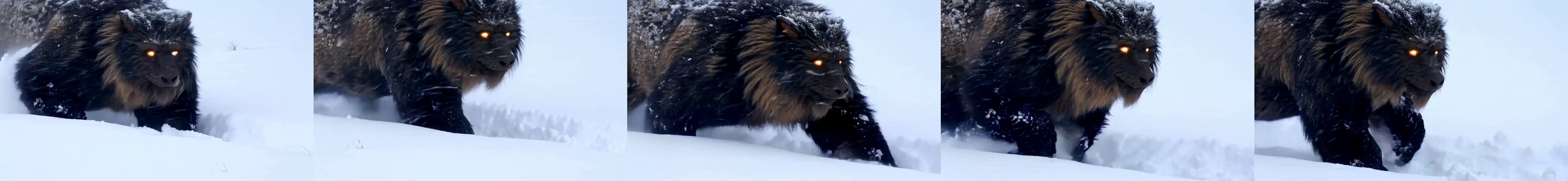} \\
        [\rowsqueeze]

        \makebox[0pt][r]{\raisebox{0pt}[0pt][0pt]{\rotatebox{90}{\scriptsize\hspace{-10pt}15$\times T_c$=5}}\hspace{\labelgap}} &
        \includegraphics[width=\linewidth]{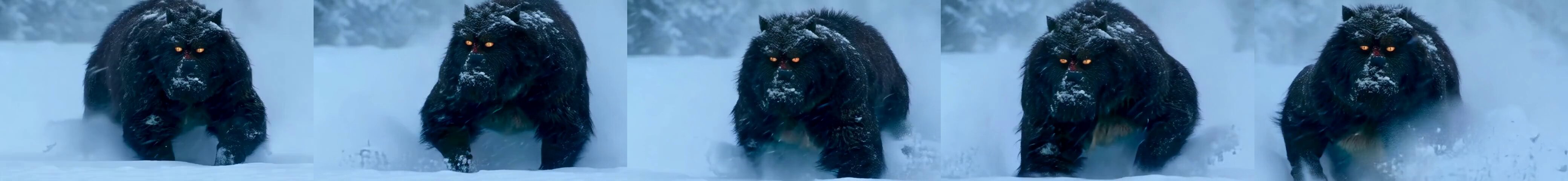} \\
        [\rowsqueeze]

        \makebox[0pt][r]{\raisebox{0pt}[0pt][0pt]{\rotatebox{90}{\scriptsize\hspace{-10pt}15$\times T_c$=3}}\hspace{\labelgap}} &
        \includegraphics[width=\linewidth]{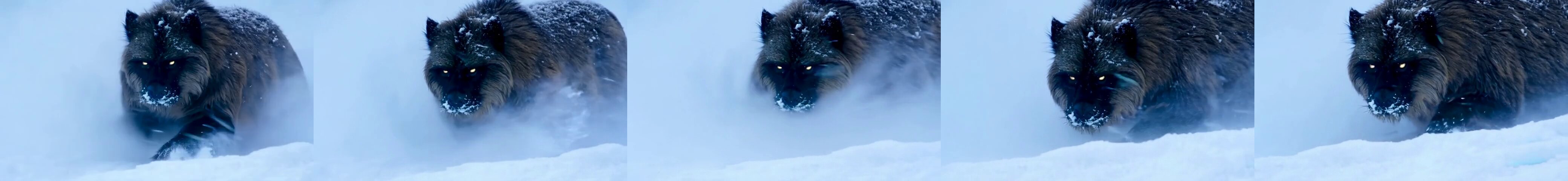} \\
        [\rowsqueeze]

        \makebox[0pt][r]{\raisebox{0pt}[0pt][0pt]{\rotatebox{90}{\scriptsize\hspace{-10pt}20$\times T_c$=3}}\hspace{\labelgap}} &
        \includegraphics[width=\linewidth]{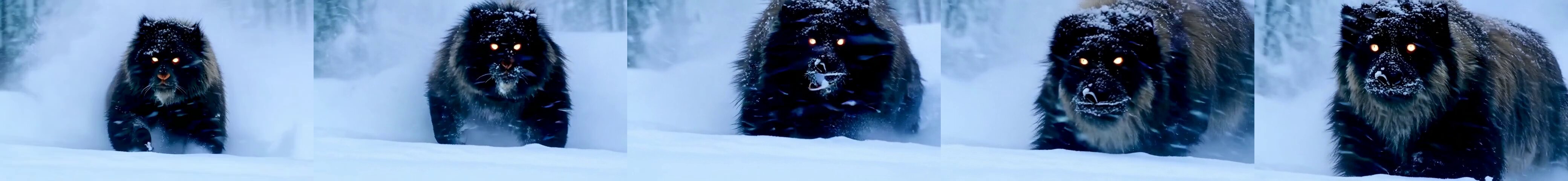} \\
        [\rowsqueeze]

    \end{tabular}
    \caption{Qualitative videos comparing original Wan2.1 1.3B model to our various hybrid variations for input prompt \emph{A bigfoot walking in the snowstorm.}}%
    \label{app:qual09}
\end{figure*}

\begin{figure*}[htbp]
    \centering
    \setlength{\tabcolsep}{0pt}
    \renewcommand{\arraystretch}{0.1}
    \begin{tabular}{@{}m{0pt}@{}m{\linewidth}@{}}
        \makebox[0pt][r]{\raisebox{0pt}[0pt][0pt]{\rotatebox{90}{\scriptsize\hspace{-20pt}Wan2.1 1.3B}}\hspace{\labelgap}} &
        \includegraphics[width=\linewidth]{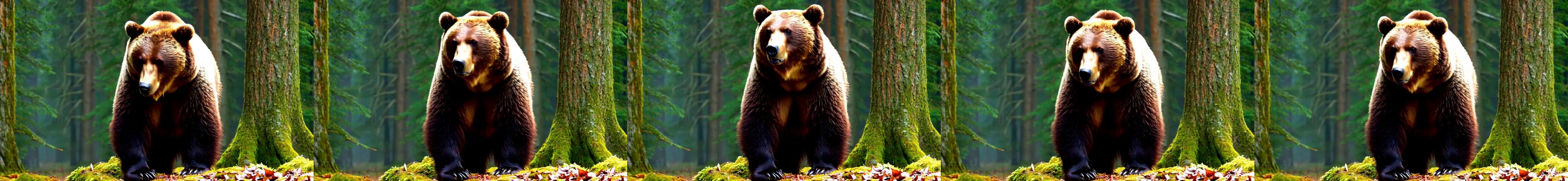} \\
        [\rowsqueeze]

        \makebox[0pt][r]{\raisebox{0pt}[0pt][0pt]{\rotatebox{90}{\scriptsize\hspace{-10pt}15$\times T_c$=5}}\hspace{\labelgap}} &
        \includegraphics[width=\linewidth]{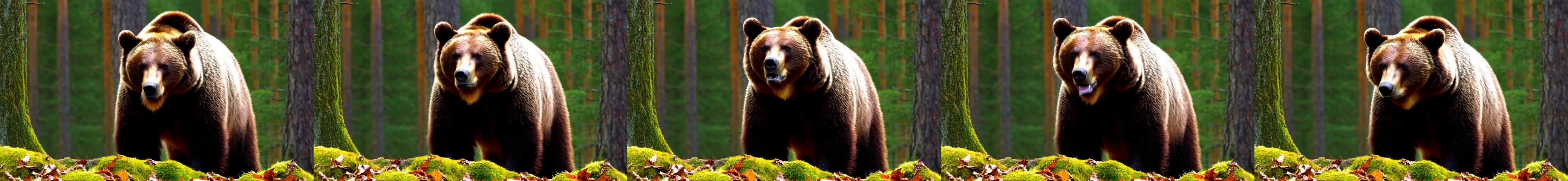} \\
        [\rowsqueeze]

        \makebox[0pt][r]{\raisebox{0pt}[0pt][0pt]{\rotatebox{90}{\scriptsize\hspace{-10pt}15$\times T_c$=3}}\hspace{\labelgap}} &
        \includegraphics[width=\linewidth]{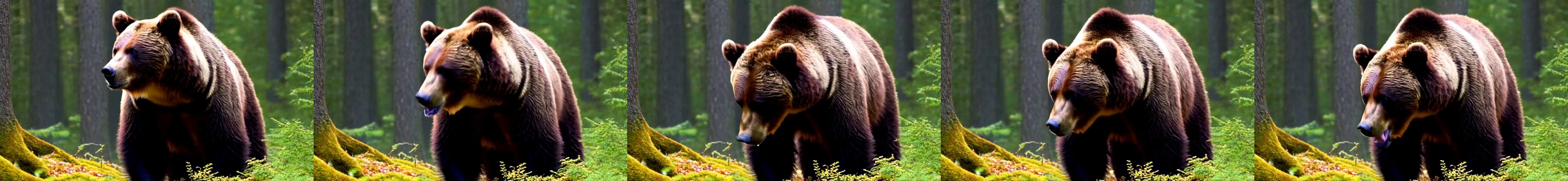} \\
        [\rowsqueeze]

        \makebox[0pt][r]{\raisebox{0pt}[0pt][0pt]{\rotatebox{90}{\scriptsize\hspace{-10pt}20$\times T_c$=3}}\hspace{\labelgap}} &
        \includegraphics[width=\linewidth]{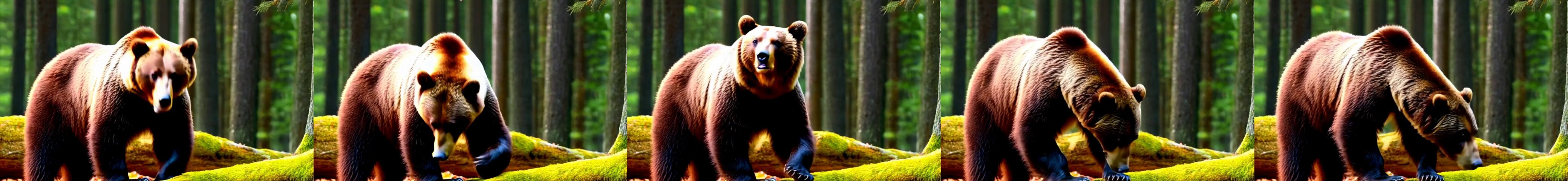} \\
        [\rowsqueeze]

    \end{tabular}
    \caption{Qualitative videos comparing original Wan2.1 1.3B model to our various hybrid variations for input prompt \emph{a bear sniffing the air for scents of food}}%
    \label{app:qual10}
\end{figure*}

\begin{figure*}[htbp]
    \centering
    \setlength{\tabcolsep}{0pt}
    \renewcommand{\arraystretch}{0.1}
    \begin{tabular}{@{}m{0pt}@{}m{\linewidth}@{}}
        \makebox[0pt][r]{\raisebox{0pt}[0pt][0pt]{\rotatebox{90}{\scriptsize\hspace{-20pt}Wan2.1 1.3B}}\hspace{\labelgap}} &
        \includegraphics[width=\linewidth]{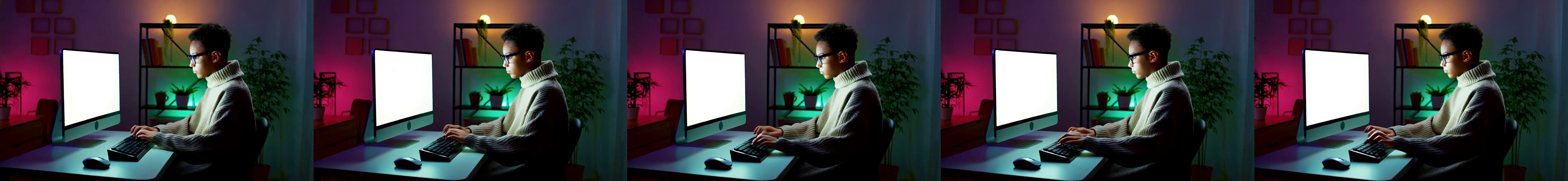} \\
        [\rowsqueeze]

        \makebox[0pt][r]{\raisebox{0pt}[0pt][0pt]{\rotatebox{90}{\scriptsize\hspace{-10pt}15$\times T_c$=5}}\hspace{\labelgap}} &
        \includegraphics[width=\linewidth]{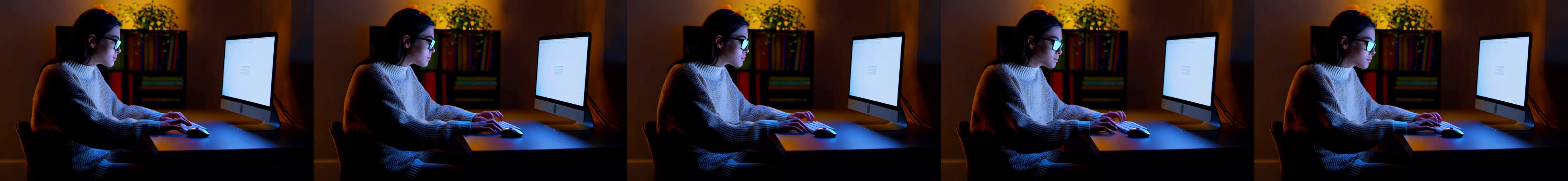} \\
        [\rowsqueeze]

        \makebox[0pt][r]{\raisebox{0pt}[0pt][0pt]{\rotatebox{90}{\scriptsize\hspace{-10pt}15$\times T_c$=3}}\hspace{\labelgap}} &
        \includegraphics[width=\linewidth]{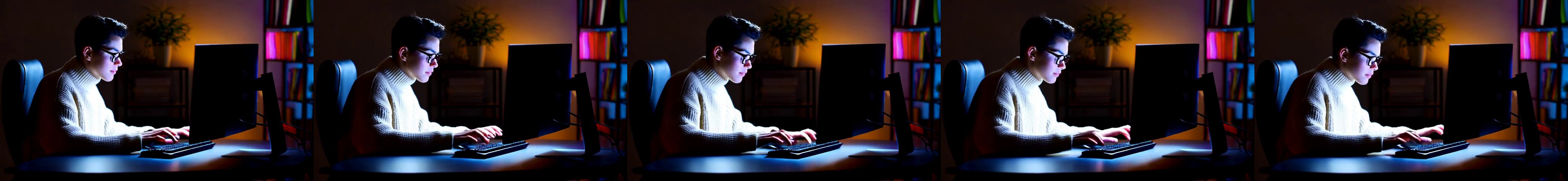} \\
        [\rowsqueeze]

        \makebox[0pt][r]{\raisebox{0pt}[0pt][0pt]{\rotatebox{90}{\scriptsize\hspace{-10pt}20$\times T_c$=3}}\hspace{\labelgap}} &
        \includegraphics[width=\linewidth]{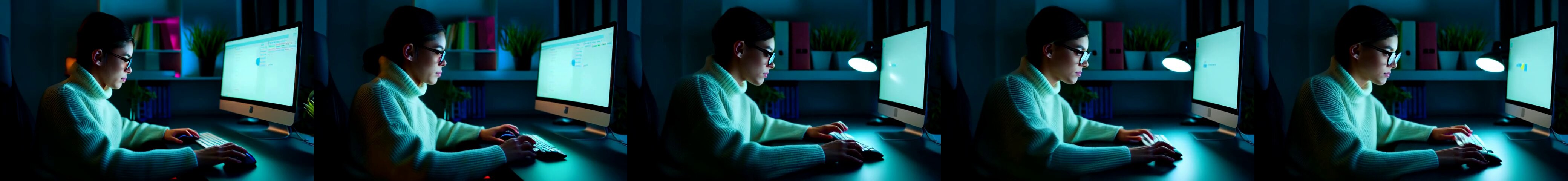} \\
        [\rowsqueeze]

    \end{tabular}
    \caption{Qualitative videos comparing original Wan2.1 1.3B model to our various hybrid variations for input prompt \emph{A person is using computer}}%
    \label{app:qual11}
\end{figure*}

\begin{figure*}[htbp]
    \centering
    \setlength{\tabcolsep}{0pt}
    \renewcommand{\arraystretch}{0.1}
    \begin{tabular}{@{}m{0pt}@{}m{\linewidth}@{}}
        \makebox[0pt][r]{\raisebox{0pt}[0pt][0pt]{\rotatebox{90}{\scriptsize\hspace{-20pt}Wan2.1 1.3B}}\hspace{\labelgap}} &
        \includegraphics[width=\linewidth]{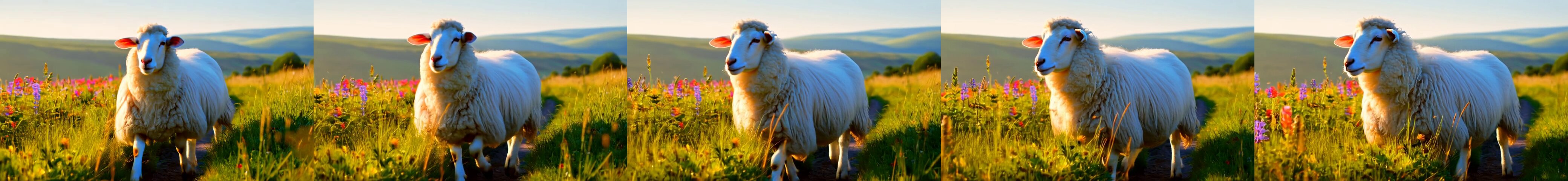} \\
        [\rowsqueeze]

        \makebox[0pt][r]{\raisebox{0pt}[0pt][0pt]{\rotatebox{90}{\scriptsize\hspace{-10pt}15$\times T_c$=5}}\hspace{\labelgap}} &
        \includegraphics[width=\linewidth]{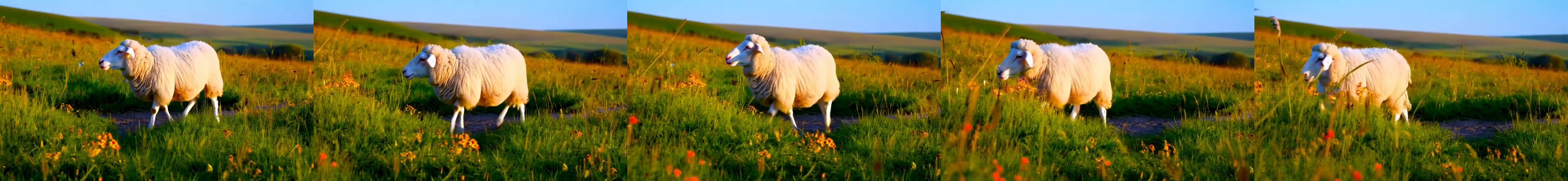} \\
        [\rowsqueeze]

        \makebox[0pt][r]{\raisebox{0pt}[0pt][0pt]{\rotatebox{90}{\scriptsize\hspace{-10pt}15$\times T_c$=3}}\hspace{\labelgap}} &
        \includegraphics[width=\linewidth]{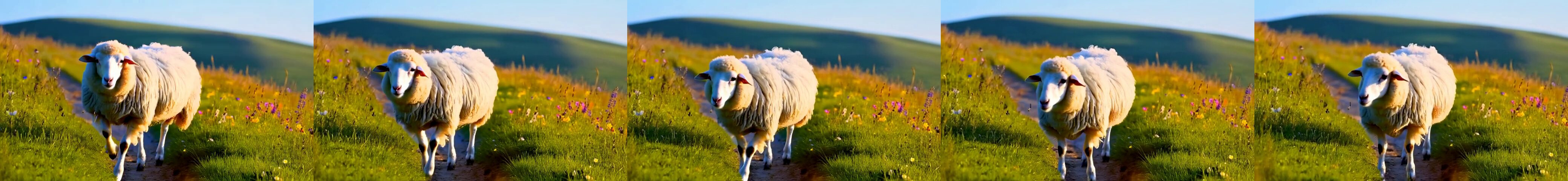} \\
        [\rowsqueeze]

        \makebox[0pt][r]{\raisebox{0pt}[0pt][0pt]{\rotatebox{90}{\scriptsize\hspace{-10pt}20$\times T_c$=3}}\hspace{\labelgap}} &
        \includegraphics[width=\linewidth]{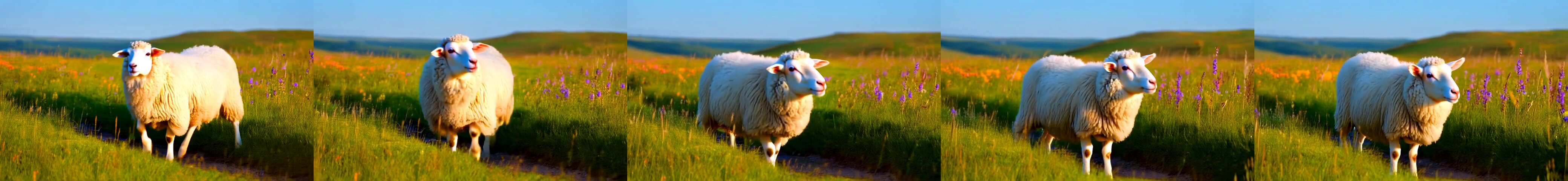} \\
        [\rowsqueeze]

    \end{tabular}
    \caption{Qualitative videos comparing original Wan2.1 1.3B model to our various hybrid variations for input prompt \emph{a sheep taking a peaceful walk}}%
    \label{app:qual12}
\end{figure*}

\begin{figure*}[htbp]
    \centering
    \setlength{\tabcolsep}{0pt}
    \renewcommand{\arraystretch}{0.1}
    \begin{tabular}{@{}m{0pt}@{}m{\linewidth}@{}}
        \makebox[0pt][r]{\raisebox{0pt}[0pt][0pt]{\rotatebox{90}{\scriptsize\hspace{-20pt}Wan2.1 1.3B}}\hspace{\labelgap}} &
        \includegraphics[width=\linewidth]{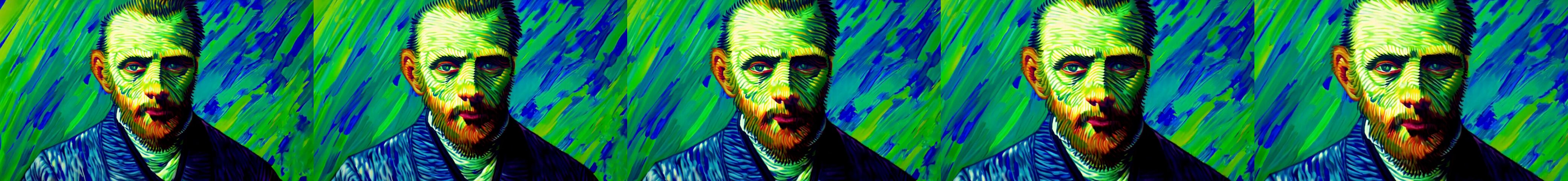} \\
        [\rowsqueeze]

        \makebox[0pt][r]{\raisebox{0pt}[0pt][0pt]{\rotatebox{90}{\scriptsize\hspace{-10pt}15$\times T_c$=5}}\hspace{\labelgap}} &
        \includegraphics[width=\linewidth]{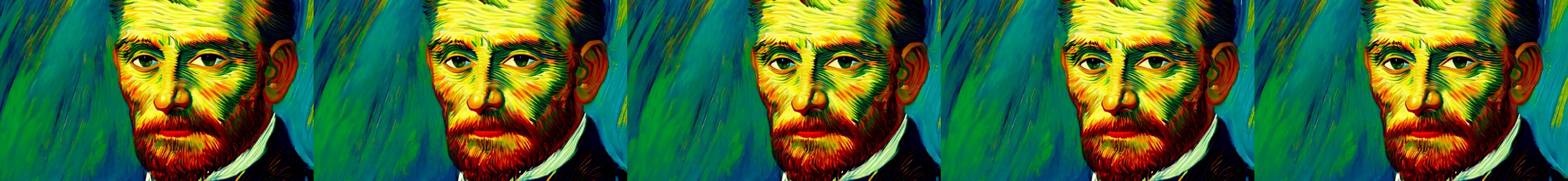} \\
        [\rowsqueeze]

        \makebox[0pt][r]{\raisebox{0pt}[0pt][0pt]{\rotatebox{90}{\scriptsize\hspace{-10pt}15$\times T_c$=3}}\hspace{\labelgap}} &
        \includegraphics[width=\linewidth]{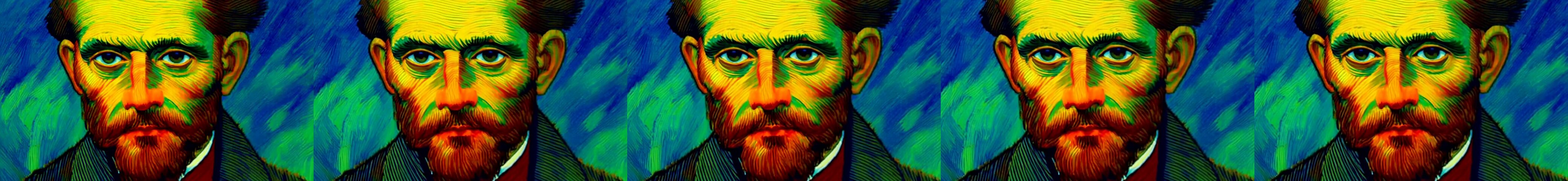} \\
        [\rowsqueeze]

        \makebox[0pt][r]{\raisebox{0pt}[0pt][0pt]{\rotatebox{90}{\scriptsize\hspace{-10pt}20$\times T_c$=3}}\hspace{\labelgap}} &
        \includegraphics[width=\linewidth]{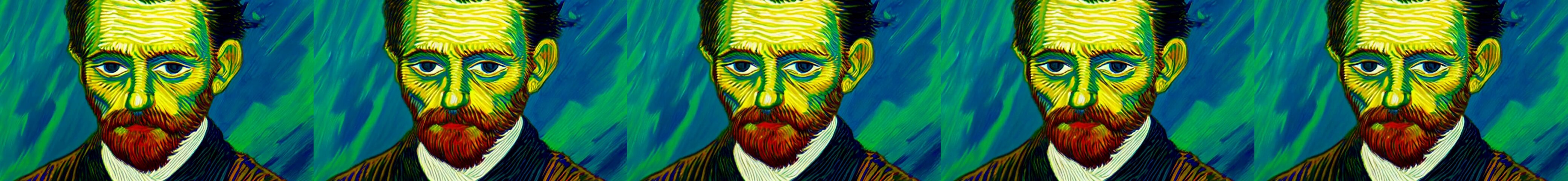} \\
        [\rowsqueeze]

    \end{tabular}
    \caption{Qualitative videos comparing original Wan2.1 1.3B model to our various hybrid variations for input prompt \emph{Cinematic shot of Van Gogh's selfie, Van Gogh style}}%
    \label{app:qual13}
\end{figure*}

\begin{figure*}[htbp]
    \centering
    \setlength{\tabcolsep}{0pt}
    \renewcommand{\arraystretch}{0.1}
    \begin{tabular}{@{}m{0pt}@{}m{\linewidth}@{}}
        \makebox[0pt][r]{\raisebox{0pt}[0pt][0pt]{\rotatebox{90}{\scriptsize\hspace{-20pt}Wan2.1 1.3B}}\hspace{\labelgap}} &
        \includegraphics[width=\linewidth]{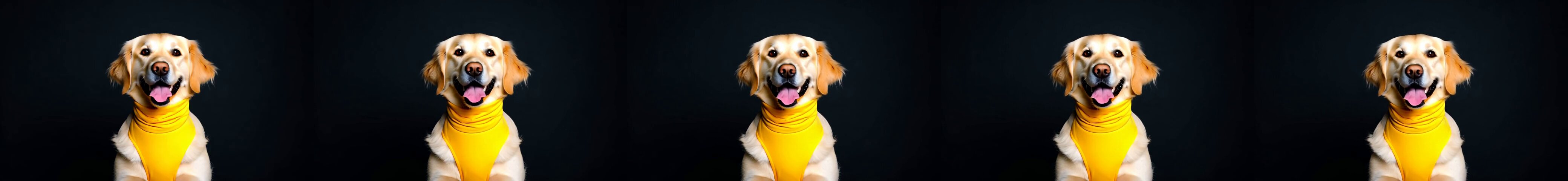} \\
        [\rowsqueeze]

        \makebox[0pt][r]{\raisebox{0pt}[0pt][0pt]{\rotatebox{90}{\scriptsize\hspace{-10pt}15$\times T_c$=5}}\hspace{\labelgap}} &
        \includegraphics[width=\linewidth]{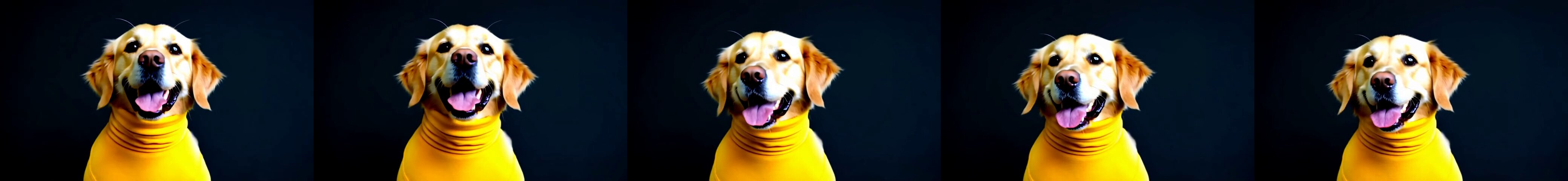} \\
        [\rowsqueeze]

        \makebox[0pt][r]{\raisebox{0pt}[0pt][0pt]{\rotatebox{90}{\scriptsize\hspace{-10pt}15$\times T_c$=3}}\hspace{\labelgap}} &
        \includegraphics[width=\linewidth]{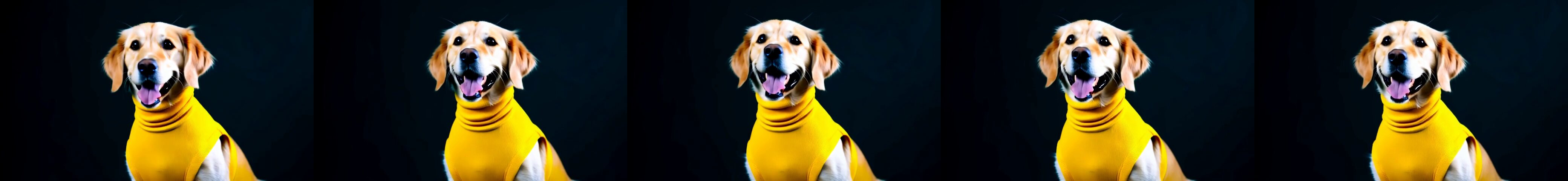} \\
        [\rowsqueeze]

        \makebox[0pt][r]{\raisebox{0pt}[0pt][0pt]{\rotatebox{90}{\scriptsize\hspace{-10pt}20$\times T_c$=3}}\hspace{\labelgap}} &
        \includegraphics[width=\linewidth]{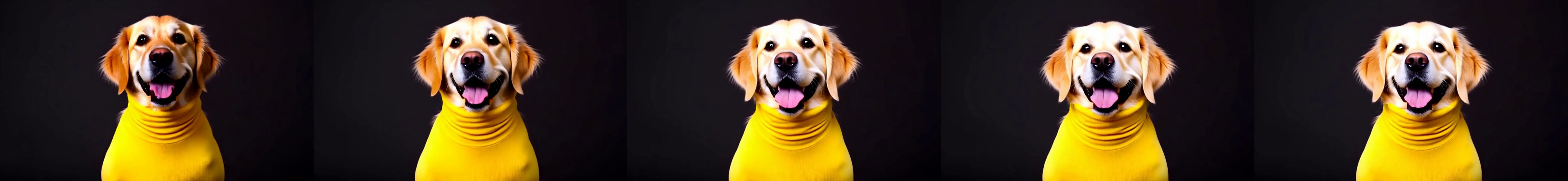} \\
        [\rowsqueeze]

    \end{tabular}
    \caption{Qualitative videos comparing original Wan2.1 1.3B model to our various hybrid variations for input prompt \emph{happy dog wearing a yellow turtleneck, studio, portrait, facing camera, dark background}}%
    \label{app:qual14}
\end{figure*}

\begin{figure*}[htbp]
    \centering
    \setlength{\tabcolsep}{0pt}
    \renewcommand{\arraystretch}{0.1}
    \begin{tabular}{@{}m{0pt}@{}m{\linewidth}@{}}
        \makebox[0pt][r]{\raisebox{0pt}[0pt][0pt]{\rotatebox{90}{\scriptsize\hspace{-20pt}Wan2.1 1.3B}}\hspace{\labelgap}} &
        \includegraphics[width=\linewidth]{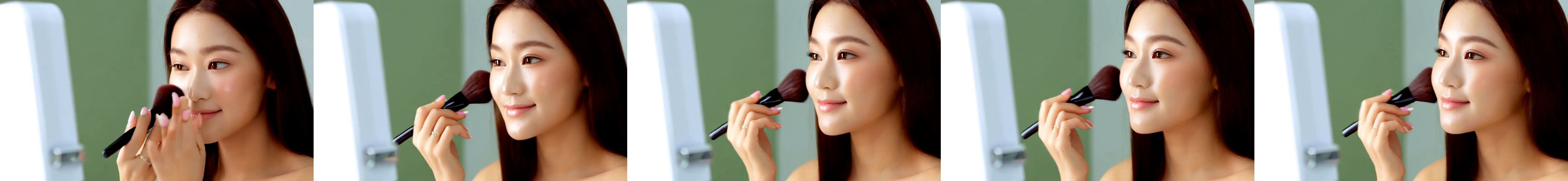} \\
        [\rowsqueeze]

        \makebox[0pt][r]{\raisebox{0pt}[0pt][0pt]{\rotatebox{90}{\scriptsize\hspace{-10pt}15$\times T_c$=5}}\hspace{\labelgap}} &
        \includegraphics[width=\linewidth]{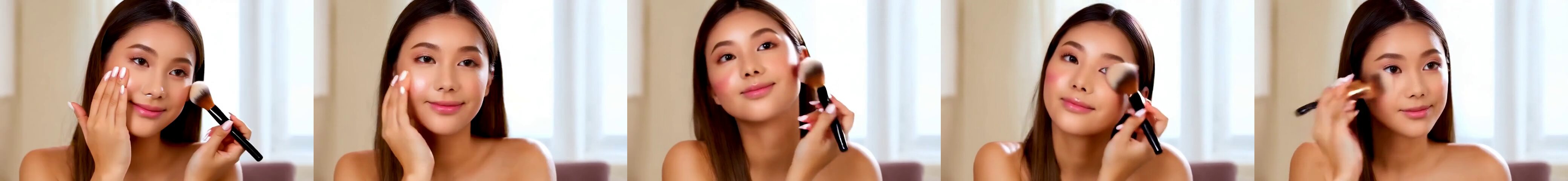} \\
        [\rowsqueeze]

        \makebox[0pt][r]{\raisebox{0pt}[0pt][0pt]{\rotatebox{90}{\scriptsize\hspace{-10pt}15$\times T_c$=3}}\hspace{\labelgap}} &
        \includegraphics[width=\linewidth]{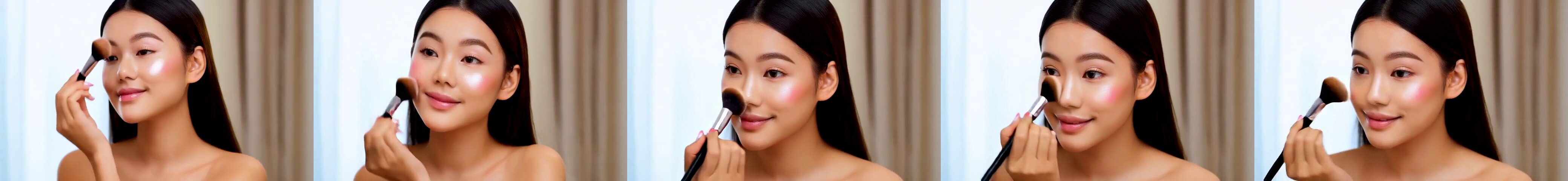} \\
        [\rowsqueeze]

        \makebox[0pt][r]{\raisebox{0pt}[0pt][0pt]{\rotatebox{90}{\scriptsize\hspace{-10pt}20$\times T_c$=3}}\hspace{\labelgap}} &
        \includegraphics[width=\linewidth]{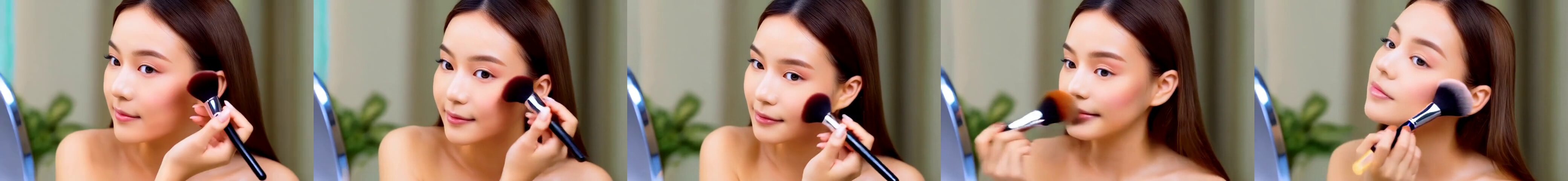} \\
        [\rowsqueeze]

    \end{tabular}
    \caption{Qualitative videos comparing original Wan2.1 1.3B model to our various hybrid variations for input prompt \emph{this is how I do makeup in the morning.}}%
    \label{app:qual15}
\end{figure*}

\begin{figure*}[htbp]
    \centering
    \setlength{\tabcolsep}{0pt}
    \renewcommand{\arraystretch}{0.1}
    \begin{tabular}{@{}m{0pt}@{}m{\linewidth}@{}}
        \makebox[0pt][r]{\raisebox{0pt}[0pt][0pt]{\rotatebox{90}{\scriptsize\hspace{-20pt}Wan2.1 1.3B}}\hspace{\labelgap}} &
        \includegraphics[width=\linewidth]{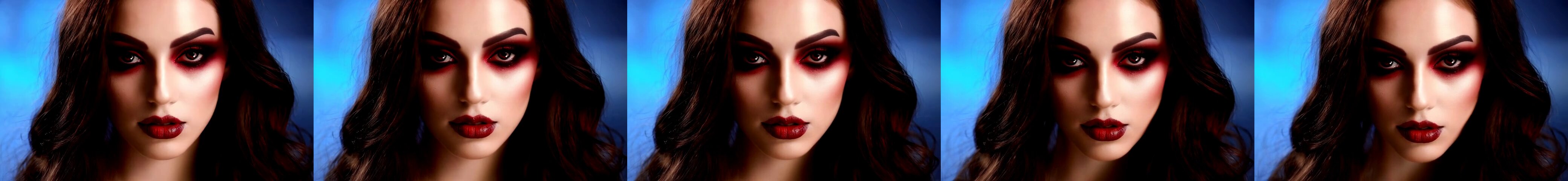} \\
        [\rowsqueeze]

        \makebox[0pt][r]{\raisebox{0pt}[0pt][0pt]{\rotatebox{90}{\scriptsize\hspace{-10pt}15$\times T_c$=5}}\hspace{\labelgap}} &
        \includegraphics[width=\linewidth]{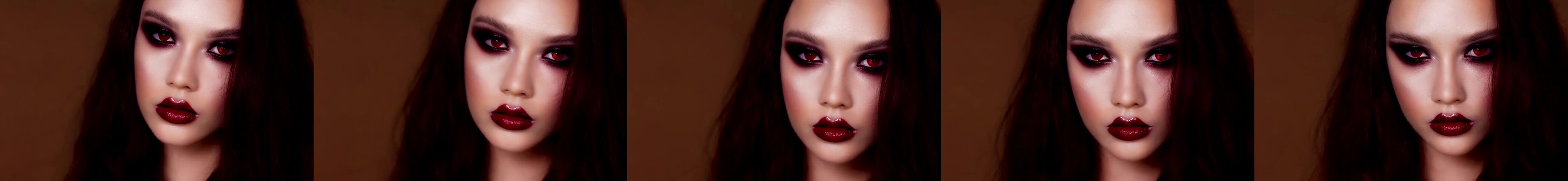} \\
        [\rowsqueeze]

        \makebox[0pt][r]{\raisebox{0pt}[0pt][0pt]{\rotatebox{90}{\scriptsize\hspace{-10pt}15$\times T_c$=3}}\hspace{\labelgap}} &
        \includegraphics[width=\linewidth]{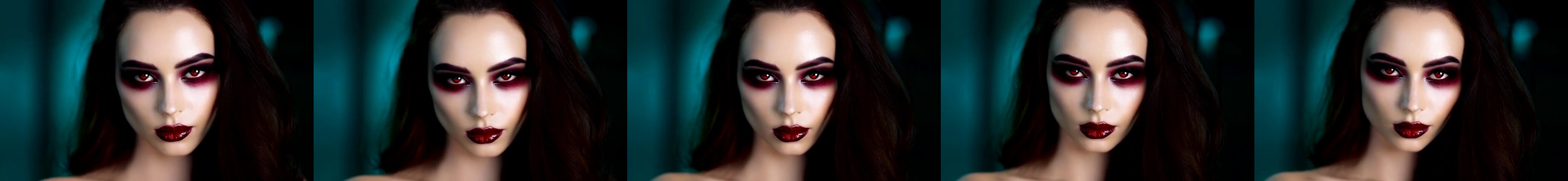} \\
        [\rowsqueeze]

        \makebox[0pt][r]{\raisebox{0pt}[0pt][0pt]{\rotatebox{90}{\scriptsize\hspace{-10pt}20$\times T_c$=3}}\hspace{\labelgap}} &
        \includegraphics[width=\linewidth]{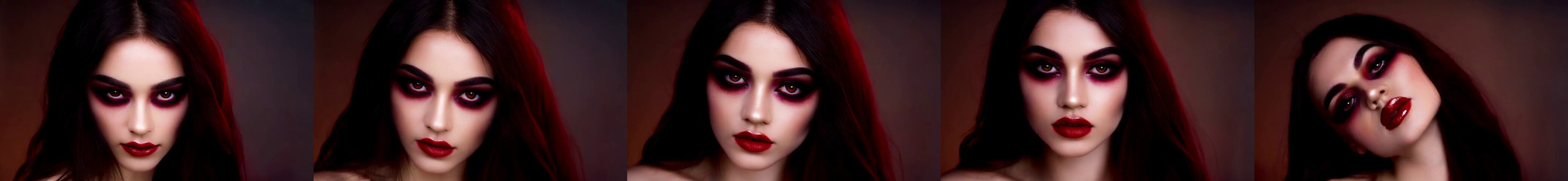} \\
        [\rowsqueeze]

    \end{tabular}
    \caption{Qualitative videos comparing original Wan2.1 1.3B model to our various hybrid variations for input prompt \emph{Vampire makeup face of beautiful girl, red contact lenses.}}%
    \label{app:qual16}
\end{figure*}

\begin{figure*}[htbp]
    \centering
    \setlength{\tabcolsep}{0pt}
    \renewcommand{\arraystretch}{0.1}
    \begin{tabular}{@{}m{0pt}@{}m{\linewidth}@{}}
        \makebox[0pt][r]{\raisebox{0pt}[0pt][0pt]{\rotatebox{90}{\scriptsize\hspace{-20pt}Wan2.1 1.3B}}\hspace{\labelgap}} &
        \includegraphics[width=\linewidth]{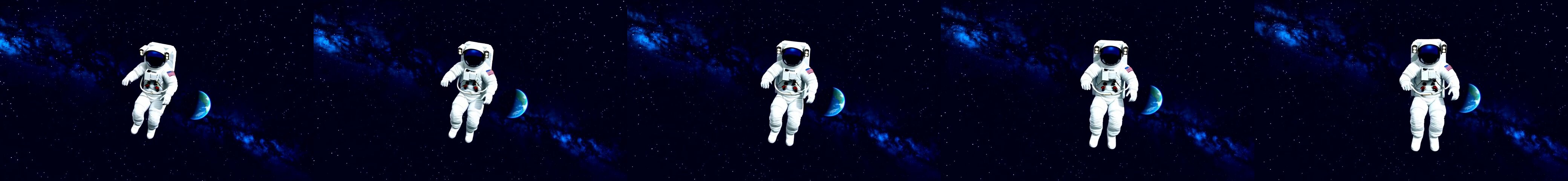} \\
        [\rowsqueeze]

        \makebox[0pt][r]{\raisebox{0pt}[0pt][0pt]{\rotatebox{90}{\scriptsize\hspace{-10pt}15$\times T_c$=5}}\hspace{\labelgap}} &
        \includegraphics[width=\linewidth]{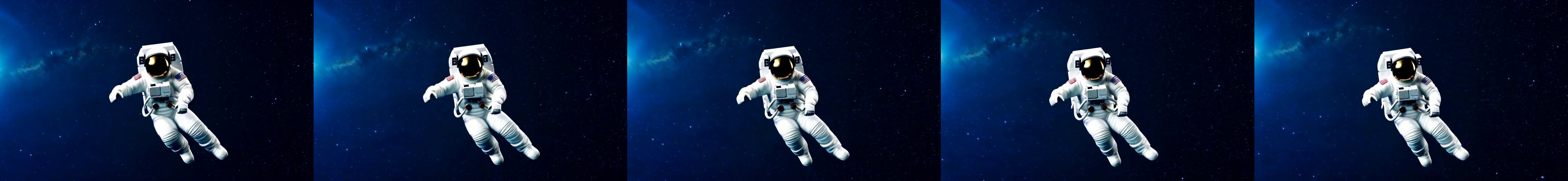} \\
        [\rowsqueeze]

        \makebox[0pt][r]{\raisebox{0pt}[0pt][0pt]{\rotatebox{90}{\scriptsize\hspace{-10pt}15$\times T_c$=3}}\hspace{\labelgap}} &
        \includegraphics[width=\linewidth]{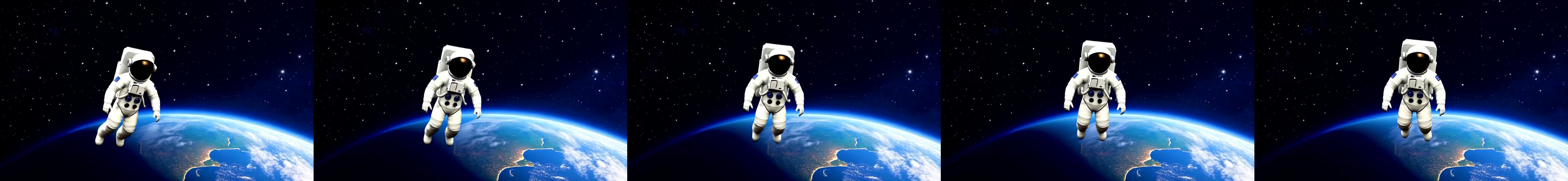} \\
        [\rowsqueeze]

        \makebox[0pt][r]{\raisebox{0pt}[0pt][0pt]{\rotatebox{90}{\scriptsize\hspace{-10pt}20$\times T_c$=3}}\hspace{\labelgap}} &
        \includegraphics[width=\linewidth]{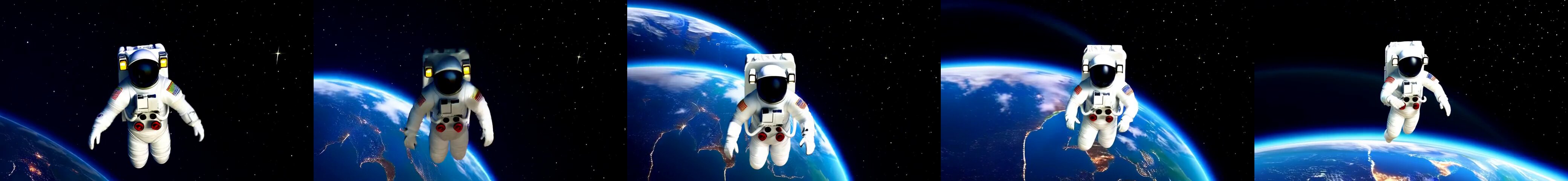} \\
        [\rowsqueeze]

    \end{tabular}
    \caption{Qualitative videos comparing original Wan2.1 1.3B model to our various hybrid variations for input prompt \emph{An astronaut flying in space, featuring a steady and smooth perspective}}%
    \label{app:qual17}

\end{figure*}

\end{document}